\documentclass[12pt]{article}

\usepackage{scicite}
\usepackage{times}
\usepackage{amsmath,amssymb,amsopn,amstext,amsfonts, amsthm}
\usepackage[pdftex]{graphicx}
\usepackage{hyperref}
\usepackage[dvipsnames]{xcolor}
\usepackage{url}
\DeclareUnicodeCharacter{0308}{\"}

\usepackage{tikz}

\usepackage{xcolor}

\newenvironment{sciabstract}{%
	\begin{quote} \bf}
	{\end{quote}}

\title{Microsaccade-inspired Event Camera for Robotics}

\author
{Botao He*$^{1,2,{\dag}}$, Ze Wang$^{3,4}$, Yuan Zhou$^{2,3}$, Jingxi Chen$^{1}$, Chahat Deep Singh$^{1}$, \\Haojia Li$^{5}$, Yuman Gao$^{2,3}$, Shaojie Shen$^{5}$, Kaiwei Wang$^{4}$, Yanjun Cao$^{3}$, \\Chao Xu$^{2,3}$, Yiannis Aloimonos$^{1,6,7}$, Fei Gao*$^{2,3}$, and Cornelia Ferm\"uller*$^{1,6,7}$\\
\\
	\normalsize{$^{1}$Department of Computer Science,}\\
        \normalsize{University of Maryland, College Park, MD 20742, USA. }\\
	\normalsize{$^{2}$College of Control Science and Engineering,}\\
	\normalsize{Zhejiang University, Hangzhou, China.}\\
	\normalsize{$^{3}$Huzhou Institute of Zhejiang University, Huzhou, China.}\\
        \normalsize{$^{4}$College of Optical Science and Engineering,}\\
        \normalsize{Zhejiang University, Hangzhou, China.}\\
        \normalsize{$^{5}$Department of Electronic and Computer Engineering,}\\
        \normalsize{Hong Kong University of Science and Technology, Hong Kong, China. } \\
        \normalsize{$^{6}$Institute for Advance Computer Studies,}\\
        \normalsize{University of Maryland, College Park, MD 20742, USA. }\\
        \normalsize{$^{7}$Institute for Systems Research,}\\
        \normalsize{University of Maryland, College Park, MD 20742, USA. }\\
	\\
        \small{Project Website: https://bottle101.github.io/AMI-EV/} \\
        \\
	\normalsize{$^\ast$ Corresponding author. Email: fgaoaa@zju.edu.cn (Fei Gao);}\\
        \normalsize{fermulcm@umd.edu (Cornelia Ferm\"uller); botao@umd.edu (Botao He). }
}
\date{}

\begin{document}
    \clearpage


	\baselineskip24pt


	\maketitle

\let\thefootnote\relax\footnotetext{${\dag}$ Part of the work was done when Botao He was a member of Zhejiang University. Original idea raised at Zhejiang University.}

\begin{sciabstract} %in microseconds
    Neuromorphic vision sensors or event cameras have made the visual perception of extremely low reaction time possible, opening new avenues for high-dynamic robotics applications. These event cameras' output is dependent on both motion and texture. However, the event camera fails to capture object edges that are parallel to the camera motion. 
    % However, their outputs are dependent on both motion and texture with edges parallel to the image motion not visible and textures fading away when the motion becomes small.
    This is a problem intrinsic to the sensor and therefore challenging to solve algorithmically.
    % Human vision employs small eye movements to overcome perceptual fading,
    % %during fixation
    % with the most prominent ones called the
    Human vision deals with perceptual fading using the active mechanism of small involuntary eye movements -- the most prominent ones
    %also has  to deal with perceptual fading, and to overcome it humans have unique active perception mechanisms, with the most prominent ones 
    called microsaccades.
    By moving the eyes constantly and slightly during fixation, microsaccades can substantially maintain texture stability and persistence.
    Inspired by microsaccades, we designed an event-based perception system capable of simultaneously maintaining low reaction time and stable texture.
    In this design, a rotating wedge prism was mounted in front of the aperture of an event camera 
    %aperture 
    to redirect light and trigger events.
    The geometrical optics of the rotating wedge prism allows for algorithmic compensation of the additional rotational motion, resulting in a stable texture appearance and high informational output independent of external motion.
    The hardware device and software solution are integrated into 
    %the developed 
    a system, which we call Artificial MIcrosaccade-enhanced EVent camera (AMI-EV).
    Benchmark comparisons validate the superior 
    data quality of AMI-EV recordings in scenarios where both standard cameras and event cameras fail to deliver.
    %performance of AMI-EV in data quality. 
    Various real-world experiments demonstrate the potential of the system to facilitate 
    %our proposed system in facilitating
    robotics perception 
    both for low-level and high-level vision tasks. 
\end{sciabstract}

	% In setting up this template for *Science* papers, we've used both
	% the \section** command and the \paragraph* command for topical
	% divisions.  Which you use will of course depend on the type of paper
	% you're writing.  Review Articles tend to have displayed headings, for
	% which \section** is more appropriate; Research Articles, when they have
	% formal topical divisions at all, tend to signal them with bold text
	% that runs into the paragraph, for which \paragraph* is the right
	% choice.  Either way, use the asterisk (*) modifier, as shown, to

    \section*{Summary}
    % A fully autonomous swarm composed of palm-sized intelligent drones with versatile task extensibility in the wild is proposed.
    % We proposed an AMI-EV system that outperforms both standard cameras and event cameras at low-level vision tasks like feature detection and high-level vision tasks such as human pose estimation, especially in challenging scenarios, such as high frame rate and low light.
    An artificial microsaccade-enhanced event camera for varied vision tasks in challenging scenarios is proposed.

    \section*{Introduction}
    \label{sec:Introduction}
	
	Humans still outperform the most advanced robots in visual perception. Our visual systems have evolved over millions of years to help us efficiently obtain the information necessary to act in our environments.
 % Their visual systems appear to be much more effective in dealing with real-world tasks than their artificial counterparts. 
A characteristic of human vision are fixational eye movements, which are small, involuntary displacements of the eyeball. The largest of these eye movements are called microsaccades \cite{martinez2004role}.
 %\cite{wiki:microsaccade}. 
 %   A microsaccade is a tiny involuntary eye movement, which typically occurs during visual fixation. Although tiny and unnoticeable, it plays an indispensable role in human visual perception. 
    % Without microsaccades, humans cannot clearly see static objects, as is demonstrated in Fig. \ref{fig:ms} and Movie 1. Readers can experience microsaccades by following the instructions provided in the caption of Fig. \ref{fig:ms}.
    They ensure that vision does not fade during fixations \cite{Alexander2019} by generating movement and stimuli in visual neurons and also enhancing perception of spatial detail \cite{rucci2007miniature}.
    Without microsaccades, humans cannot maintain the perception of static objects. For a demonstration, see Fig. \ref{fig:ms}, and Movie 1. %Microsccades can also enhance the perception of spatial details\cite{rucci2007miniature}, but their complete functional role is still debated.
    %clearly see static objects, as for whom following the instructions in Fig. \ref{fig:ms}. A video explaining this phenomenon is Movie S1.
% 	This kind of micro action, called Microsaccade, are tiny involuntary fixational eye movements.
    The question we ask here is, can we adopt this active perception mechanism in robot vision?
    %what can this tiny active perception mechanism do for modern robotics? 
    
    % Microsaccade: https://en.wikipedia.org/wiki/Microsaccade
    % Ocular Termor and Micro Tremor: https://en.wikipedia.org/wiki/Ocular_tremor
 %   To figure it out, we first need to understand how microsaccade works.  
 %   The retina \cite{wiki:retina} is especially sensitive to high-frequency spatial information and illumination change. When there is not much photometric change in the receptive field, humans could use microsaccades to actively change the illumination of retinal images. By continuously generating stimuli in visual neurons \cite{martinez2004role}, microsaccades can substantially enhance perception of spatial details\cite{rucci2007miniature} and prevent the retinal image from fading away\cite{Alexander2019}.
    
    % Nature has been a source of inspiration for engineers since a long time. Biological systems have long served as the first and best models for engineering. Over the past century, humans have made every effort to imitate biological visual systems.
    % Microsaccades are tiny involuntary fixational eye movements. By continuously generate stimuli in visual neurons [ref?], microsaccades can substantially enhance perception of spatial details [ref] and prevent the retinal image from fading away [ref]. Many researches demonstrate the remarkable capability of Microsaccades in \textit{fixation correction, memory,[ref.] control of binocular fixation disparity[ref.] and attentional shifts.[ref.]}

 %   In robotics, there is a
    A bio-inspired visual motion sensor, known as the silicon retina, Dynamic Vision Sensor (DVS) \cite{lenero20113}, or event camera, has recently gained increasing attention in robotics. 
    %By mimicking the retina, 
    Using analog microcircuits at every pixel,
    it can achieve a temporal resolution of several microseconds and has much higher dynamic range than standard cameras.
    %, which has greatly facilitated its use in visual perception.
  %  In the past few years, the event camera has gotten more and more attention. The number of publications in this field reached 1339 during 2021\cite{dimentions}, which is $196\%$ than it was in 2016, and the rate is still rapidly increasing. 
 % Moreover, 
 Event cameras have  shown great potential in many visual navigation tasks, including dynamic obstacle sensing~\cite{falanga2020dynamic, he2021fast, mitrokhin2020learning, sanket2019evdodge}, localization in challenging lighting conditions\cite{zhou2021event, rebecq2016evo, vidal2018ultimate,mitrokhin2018event}, and specific applications such as autonomous inspection\cite{dietsche2021powerline} or space situational awareness \cite{cohen2019event}. 
%   https://app.dimensions.ai/discover/publication?search_mode=content&search_text=event%20camera&search_type=kws&search_field=text_search
    However, along with these functional advantages, some of its natural properties also present unique challenges.

    % \begin{figure}[htb]
    % \centering
    % \includegraphics[width=1\textwidth]{figures/fig2.png}
    % \caption{Illustration for losing shape.}
    % \label{fig:illu_shape}
    % \end{figure}

   Event cameras only respond to motion. An event at a pixel is triggered when the logarithm of the intensity changes by a certain threshold. Thus the readings occur at image edges, but depend on both the motion and the scene texture.  
   % One fundamental property of event cameras is that they only respond to motion (assuming constant illumination). Motion constantly evolves over time, with the same motion pattern typically lasting for a brief period, and changes in the pattern result in alterations to the texture's appearance. In a nutshell, events are not only dependent on the scene but also dependent on the camera movement. 
   % For intuition, see Fig. \ref{fig:texture}B, upper row. 
   No events are recorded at edges parallel to the camera motion, and thus an event camera moving horizontally does not "see" horizontal scene edges. 
    As a result, event cameras do not produce a stable and persistent texture, and they cannot maintain high informational output all the time, which makes accurate and long-term data association very difficult. 
    However, data association 
    %is the most fundamental and essential part of 
    is essential for most algorithms employed in robot visual perception systems, such as optical flow estimation or feature tracking.
    %As a result, this issue – the dependence of the events on the motion and not just the scene -  has 
    The challenge of maintaining it has become a bottleneck 
    for event-based vision in real-world applications. 
   % that limits the development of event-based visual perception.

    % 直接跟踪：FAST, Arc*; optical flow. 运动模式突变
    
    % maintain map: EVO。 信噪比降低
    % 重建后跟踪：传统方法: static ok, but noise will make it fail after seconds, or sacrifice HDR, e2vid改善但是指标不治本。。。。noise---在maintain map里面一起讲
     
    % fusion: 和普通相机融合，ultimate slam, EDS。失去HDR等优势    
    
    In the past decade, many works attempted to eliminate this problem using software approaches.
    Most event-based data association methods rely on features like corner points\cite{mueggler2017fast, alzugaray2018asynchronous, GWPHKU:EVIO} and optical flow\cite{ye2018unsupervised, paredes2019unsupervised, hagenaars2021self}. 
    However, because of the varying texture appearance, 
    %their application in robotics is still scarce because the texture change also makes 
    feature detection and tracking are not accurate and stable and so far there are very few 
    %so far with limited 
    robotics applications.
    %as accurate and stable as needed. 
    In recent years, some works \cite{messikommer2023data, mitrokhin2018event, GWPHKU:EVIO, paredes2021back} associated
    events with previous data maintained either in the form of 2D/3D event maps or reconstructed intensity images, and optimized the correspondence between new and  maintained data.
    The maintained maps or images contain more information and have enhanced texture stability, thus resulting in more robust performance.
    %compared with matching events directly.
    However, these methods suffer from noise when the event camera moves slowly or is static, resulting in severe robustness issues if such conditions persist over extended time intervals.
 Some works combined event sensors with regular cameras for optical flow estimation \cite{barranco2014contour} and stable feature tracking \cite{hidalgo2022event,PL-EVIO, ESVIO}.
%    Recently, Hidalgo-Carrió et al.\cite{hidalgo2022event} and Guan et al.\cite{PL-EVIO, ESVIO} combine event cameras with regular cameras to utilize intensity images for stable feature tracking. 
By fusing events with absolute brightness information, 
%these methods can detect features from regular intensity images and track them with events. 
features can be detected in the intensity images and  tracked with events. However, the introduction of regular cameras limits the system's dynamic range, thus hindering its application in challenging lighting environments.
    % leads to many robustness issues due to its poor performance in HDR, motion blur and dark scenarios.
    All the above methods attempt to maintain a stable texture appearance
    using software solutions. Although they offer some mitigation, they fall short of providing a complete solution.
    %Stable data association is still an unsolved problem in event-based vision.
    %combined with standard hardware.  
   % from the software perspective. 
    We observed that the issues of texture instability and information loss are fundamentally introduced by sensor characteristics instead of algorithm imperfections. 

In recent years, people tried to address this problem via active vision approaches. Several studies have integrated event cameras with other active sensors, such as structured light or lasers \cite{brandli2014adaptive, matsuda2015mc3d, muglikar2021esl, muglikar2023event, muglikar2021eventevdepth}, to facilitate motion-independent event sensing. These studies introduce specialized sensor configurations that demonstrate impressive results in tasks like depth estimation, 3D reconstruction, and surface normal estimation, the unique setups limit their adaptability to diverse applications. Moreover, these configurations tend to be more susceptible to specific illumination conditions and material types, constraining their broader utility.
 Some previous works emulated the human microsaccade mechanism
by introducing additional motion into the event camera system\cite{orchard2015converting, yousefzadeh2018active}. 
%In these works, 
By shaking the event camera and introducing movements in different directions using a pan-tilt mechanism, saccade-like motions were introduced, and more information (events) could be recorded from multiple saccades. 
%{\it The argument is hard to understand}
   % However, these methods cannot handle dynamic scenarios (the presence of independently moving objects). This is because to acquire several saccades describing the same environment, the image scene cannot change much during a period of saccades, which typically takes $500ms$. This constraint highly limits the move-ability of the system and is thus not scalable for general robotics applications.
  %  To eliminate the movability problem, one period of saccades should be finished as soon as possible, and to achieve this, simply mimicking nature may not be enough.
    % This is because the electronic perception system has large inertia and to make it vibrate at high frequencies requires large torque, which is difficult to achieve with the existing light-weighted actuators.
    However, discrete sensor movements are difficult to implement in robotics systems. 
    This is due to the substantial inertia of the electronic perception system; achieving high-frequency vibrations necessitates considerable torque, which is challenging to accomplish using currently available lightweight actuators.
    Therefore, to effectively address the issue of fading, alternative approaches inspired by nature, rather than strictly mimicking it, are required.

 %   \textcolor{red}{ I don't think that saccades adjust the direction of the incoming light. This paragraph should be rewritten.}
   % The primary aim of incorporating active motion is to adjust the direction of the incoming light. 
   % This infers that if we can redirect the incoming light without necessitating the movement of the cumbersome perception system, a substantial reduction in the required torque can be achieved. 
    Our aim is to develop a similar Artificial Microsaccade (AMI) mechanism that vary the direction between the scene texture and the image motion. Although this can be done with saccades, it can also be achieved by manipulating the direction of the incoming light.
    Moreover, if the direction of the incoming light can be steered continuously rather than in discrete steps, the efficiency will also be improved. This is the basic idea that we use to design our system that will “see” events at all edges of the scene and will not miss any due to its motion.

    \subsection*{Proposed Solution}
   % By mimicking the microsaccade mechanism of human,
   This paper identifies and resolves fundamental challenges to achieving accurate and stable event-driven data association from the perspective of hardware-software joint design. 
   Instead of simply replicating nature, we propose a nature-inspired but more effective solution that utilize AMI mechanism to manipulate the direction of the incoming light, named Artificial MIcrosaccade-enhanced EVent camera (AMI-EV).
 %   In this work, 
 %   We design a 
 The AMI-EV actively senses visual information using a rotating wedge prism in front of an event camera. By actively triggering events in areas of high spatial frequency, such as edges,  AMI-EV maintains the appearance of texture and high informational output, even when the sensor does not move. 
 Fig.~\ref{fig:system}A illustrates the hardware,  
 %is a rotating wedge-prism camera system, called 
 %of our \textbf{A}rtificial \textbf{MI}crosaccade-enhanced \textbf{EV}ent camera (AMI-EV) 
Fig. \ref{fig:system}B the refraction of the wedge mechanism, 
 and  Fig. \ref{fig:system}C the imaging.
 %Fig. \ref{fig:system}C,
    %by generating omnidirectional microsaccades,
Details of the rotating wedge-prism mechanism and the compensation algorithm are  described in the Materials and Methods and in Suppl. Movie. S1. 
%    By actively triggering events in areas of high spatial frequency, i.e., edges,  AMI-EV maintains the appearance of texture and high informational output, even when it is static. 
   % We reveal its mechanism as shown in Fig. \ref{fig:system}B and Fig. \ref{fig:system}C, and develop an AMI compensation algorithm to make our system a plug-in-and-use solution with existing event-based perception algorithms. Details about the compensation algorithm can be found in Materials and Methods.
   The compensation algorithm makes our system a plug-in-and-use solution with existing event-based perception algorithms.
    We validate the potential of the system by applying it to many different low- and high-level vision tasks as detailed in the Results.
    To facilitate future research, we also release
    %proposed hardware-software solution, including 
    the hardware design, the software for AMI generation, calibration and compensation, a simulation platform, and a translator for interfacing with public event camera datasets. With these tools \cite{botao_he_2023_8175198}, developers can generate their own AMI-EV datasets for their specific tasks from simulation, existing event-based vision datasets, and real-world environments.
    %, all based on our platform.
    % \note{Modify: add references about previous saccade works}

    \section*{Results}
    In this section, we present the design of our AMI mechanism and then demonstrate its advantages due to its capability of maintaining stable and high informational output. 
    To 
    demonstrate %illustrate the performance and 
    the 
    %proposed 
    system's potential in facilitating robotics perception research, 
    %from different perspectives, 
   % we integrate our system into 
    we evaluated it on various state-of-the-art event-based algorithms in several typical applications. 
    %Extensive quantitative evaluations
    The results verify that the proposed system is highly effective in improving performance across the board.

    \subsection*{Artificial Microsaccade Generation and Compensation}
    To generate events on all edges, we utilized the working principle of the wedge-prism deflector\cite{senderakova2003analysis}. When the prism rotates, it actively adjusts the direction of the incoming light, as illustrated in Fig. \ref{fig:system}B. 
    At the beginning of the procedure, 
    the wedge prism has a certain orientation and deflects the incoming light at a fixed angle, as shown  in Fig. \ref{fig:system}B(i). Then the actuator module drives the optical deflector module to rotate along the Z-axis of the camera, $\mathbf{z_c}$, to make the incoming light constantly change its deflection, as shown in Fig. \ref{fig:system}B(ii).
    This enables the incoming light to continually generate events as it creates a motion on the image plane with a circle-like trajectory, as shown in Fig. \ref{fig:system}B(iii).
   % A discussion on optimal refraction angle and frequency of the rotations  for  different tasks is available in the Materials and Methods section.
    As a result, continuously changing rotational motion is induced in the camera. Since the AMI is in all directions in the image plane, the output event stream contains all boundary information of the scene, as shown in Fig. \ref{fig:system}(C and D). 
    Compared with previous works \cite{orchard2015converting, yousefzadeh2018active} that move the camera instead of the prism, the moving parts of our system do not contain fragile components such as the camera. Thus rendering it more robust for high-speed rotation. Moreover, our system operates under constant-speed rotation, which is a smoother motion than the vibrational motion considered in \cite{orchard2015converting, yousefzadeh2018active}.
    A discussion on the optimal refraction angle and frequency of the rotations for different tasks is available in the Materials and Methods section.

% 	calibration: (with ignorable error analyzed in appendix XX)	
    Another important part of the proposed software framework is the AMI compensation. This is one of the major advantages of our approach compared to previous works\cite{orchard2015converting, yousefzadeh2018active}, which inevitably suffer from motion blur and decreased accuracy. 
   % In this work, 
 %   The controlled and observable motion generated by rotating the wedge prism in our hardware design allows for compensation in software.
 Looking at an image created by binning the events over a small time interval, which we call an accumulated event image (see Fig. \ref{fig:system}C), blurred boundaries are observed in the absence of motion compensation.
 %   However, the event slices (i.e. the images created by binning the events over a small time interval) show blurry boundaries due to the camera motionof the wedge prism, as illustrated in the first row of Fig. \ref{fig:system}C.
To obtain sharp edges, events triggered by the same incoming light ray direction must be moved to the same pixel. This requires calibrating the wedge orientation and compensating for the spatial displacement of the events introduced by the wedge motion. 
Given that our actuator system is equipped with an absolute position sensor (rotary encoder), the compensation parameters only need to be calibrated once and can be used directly for subsequent recordings.
The technical details of the calibration and compensation algorithms are provided in the Materials and Methods section. The compensation is illustrated in the second row of Fig. \ref{fig:system}C and in Fig. \ref{fig:system}D, and Suppl. Movie S1 shows the procedure.
    
\subsection*{Quantitative Evaluation of Texture Enhancement}
    
 %    \begin{figure*}[htb]
	% \vspace{0.0cm}
	% \centering
	% \includegraphics[width=1.0\textwidth]{figures/texture.pdf}
	% \vspace{-0.6cm}
	% \caption{
	% 	The illustration of the improvement from our approach on texture enhancement in input event stream. \textbf{For event stream: } Fig. (a) is the original event stream without our mechanism, figure (b) is the event stream using our mechanism which is denser in event stream, figure (c) is the entropy histogram comparison of the original event stream and our enhanced event stream. \textbf{For accumulated event images:}  Fig. (d) is a mean-square error comparison of the original events and our enhanced events on the edge capture accuracy, figure (e) is the edge completeness comparison at a time point in (d), figure (f) is the entropy comparison in the accumulated event images, \textbf{For reconsructed gray-scale images:} Fig. (g) is the reconstructed gray-scale images comparison when camera is static, original event is on the left and our enhanced event is on the right, figure (h) is  is the reconstructed gray-scale images comparison when camera is moving forward, original event is on the left and our enhanced event is on the right, figure (i) is the quantitative comparison of the reconstructed image quality on SSIM and PSNR metrics.   
	% }
	% \label{fig:texture}
 %    \end{figure*}
    
    % \subsubsection*{Experiment Setup}
    \textbf{Experiment Setup} To verify the effectiveness of the proposed system in texture enhancement, we conducted experiments on three 
    %typical event 
    representations: event stream, accumulated event images, and reconstructed intensity images.
    In each experiment, the performance of our system was tested against a standard event camera (S-EV). For all cases, two motion scenarios were considered: (a) no motion and (b) motion with 6 degrees of freedom.
    All data were collected using the customized platform shown in Suppl. Fig. S1. The platform was equipped with an S-EV, an AMI-EV, and an Intel Realsense D435 camera \cite{d435} that provides Red-Green-Blue (RGB), grayscale, and depth images. The hardware framework refers to the design of \cite{9811935}. Further configuration details can be found in the Suppl. Methods.
    
    \textbf{Event stream} The event stream is a fundamental representation of event data from which all other event representations are derived.
    Therefore, enhancing the quality of the event stream can substantially improve the performance of a robotic perception system. 
    In this experiment, we aimed to demonstrate that our system can generate an event stream of higher quality, containing more environmental information, than the S-EV.
    The quality was evaluated using the point distribution, a common metric for evaluating the quality of 3D point clouds. 
    Previous works on spatial point cloud processing\cite{chen2023towards, kodors2017point} have shown that a uniform distribution of points across the environment surface is preferable, as it indicates that the point cloud has captured all the necessary data. 
    In the case of a spatial-temporal point cloud, or an event stream, the point distribution is determined by both the scene structure and the motion. However, the same metric can still be used if we apply constraints on the scene. If the scene is static and all edges have the same illumination change, the point distribution is determined only by camera motion. 
    In this scenario, a more narrow distribution means there is a higher proportion of events that share similar density. 
    This results in a more uniform event density across the event stream, thus leading to a more stable representation of scene features that is less affected by camera motion. 
    % which leads to a more uniformed event density of the event stream and thus a more stable representation of scene features independent with the camera motion.
    Therefore, the uniformity of the event stream can measure the influence of camera motion on the output.

    In our experiment, we employed Kernel Density Estimation (KDE) to compute the density of events at their locations. The variance of the KDE distribution serves as an indicator of the uniformity of the event density at the location of the events. A lower variance suggests that a greater number of events share similar density, leading to a more stable representation of scene features.
    The experiment environment contained edges oriented in various directions, with an even spatial distribution throughout.
     Fig. \ref{fig:texture}D illustrates that AMI-EV produced a more uniform point distribution than S-EV, with a variance of $0.196$ compared to $0.425$ for S-EV. This indicated that the output event stream of AMI-EV was more stable. Additionally, as detailed in Suppl. Methods, the AMI-EV data had a lower ratio of low-density components, which are more likely due to noise and provide little useful structural information. Fig. S10.
    
    \textbf{Accumulated event image} The accumulated event image is the most commonly used visualization in event-based vision tasks.  Thus, enhancing its quality will substantially improve subsequent applications that process event data in a manner akin to image processing.
    In this study, we 
    %endeavor to illustrate 
    showed that the accumulated event images produced by our system exhibit superior stability and displayed less dependence on camera motion.
    
    For this experiment, we first extracted the edges in the grayscale images using the Canny edge detector\cite{4767851}. Since the motion was small and the illumination was stable, we used them 
    %grayscale images from a standard camera can contain 
 as ground truth for the edges in the environment.
   % all edges from the environment, and thus can be utilized as ground truth. 
   Next, image registration was applied to align the images between the S-EV, AMI-EV, and ground truth.
    Finally, we measured the performance of capturing edges %perception 
    %performance using corresponding metrics.
  %  The article evaluates the quality of an event image
  using two metrics: Optimal Dataset Scale F1 (ODS-F) score and entropy, as shown in Fig. \ref{fig:texture}(A, B and E). ODS-F score is a commonly used metric for edge-detection tasks\cite{madaan2017wire, arbelaez2010contour}, whereas entropy is a widely-used parameter to quantify the amount of information present in an image. Both metrics are positively correlated with texture completeness in our experiments.
 % this case.
%    Here, 
    Referring to the figure, 
    %the S-EV 
    AMI-EV showed stable and complete recordings of edges 
    %environmental perception ability 
    when the camera is in motion. Furthermore, we see that the output from the AMI-EV shows less dependence on camera motion than that of the S-EV.
    
    In Fig. \ref{fig:texture}(A and B), our system demonstrated higher and more consistent ODS-F scores, which can be attributed to the AMI mechanism. 
    In certain motion patterns, such as the second snapshot in Fig. \ref{fig:texture}B, where the movement is parallel to most of the edges in the environment, the recordings from the S-EV can be greatly affected, whereas our system remains stable. 
    Moreover, as shown in Fig. \ref{fig:texture}B, our system produced substantial improvements in the image entropy metrics compared to S-EV, indicating that it more effectively recorded complete edge information. 
    The entropy has been calculated on the binarized event map, and only the most representative part of the result is displayed here.
    %due to the page limit. 
    For more detailed results, the reader is referred to Suppl. Fig. S8 in the Suppl. Methods.
    
    \textbf{Reconstructed intensity image} The enhancement of reconstructed intensity image quality is critical for event-based robot vision since such representation is essential in tasks like high frame rate video generation\cite{rebecq2019high, rebecq2019events}.
    In the experiment, we first reconstructed videos using the event cameras at 1000 fps, which is a typical frame rate used in high-speed imaging \cite{iwasaki2008visible, nagy2020faster}. Then, we utilized the Natural Image Quality Evaluator (NIQE)\cite{mittal2012making}, 
    which intuitively assesses how natural an image is to quantitatively evaluate the image quality.
    
    The results are shown in Fig. \ref{fig:texture}(C and F).  Fig. \ref{fig:texture}F shows the NIQE metric computed over a time interval with two-time instances ($T1$ and $T2$) highlighted, as shown in the two snapshots in Fig. \ref{fig:texture}C. At $T1$, the system is static, and at $T2$, it is moving.
    We see that both cameras show satisfying image reconstruction performance when the robot is moving (right side of Fig. \ref{fig:texture}C). The proposed method achieved better performance because it can provide more information in regions that lack camera motion, such as horizontal edges when the robot is turning. When the robot is static, the performance of the S-EV decreases due to perceptual fading, as shown in the left side of Fig. \ref{fig:texture}C. More details about perceptual fading can be found in the Perceptual fading effect in event cameras in the Suppl. Methods. On the other hand, the AMI mechanism effectively addresses the perceptual fading problem by actively providing more environmental information. Readers can refer to Suppl. Movie S2 for a more detailed illustration. In rare scenarios, the motion of the prism negates the optical flow induced by the motion of the camera, resulting in few events. In such scenarios, AMI-EV's performance degrades by a small margin. For example, at the 48th second in Suppl. Movie S2, there is a frame where the phenomenon of perceptual fading occurs, especially noticeable at the location of the ``FAST Lab'' logo on the image.

    % \subsection*{Quantitative Evaluation of Edge Perception}
    % To comprehensively verify the improvement of the proposed system in edge perception, we collect datasets with over 2000 event-edgemaps, containing all 6-DoF motions. 
    % In evaluation stage, we choose three measures: fixed contour threshold (ODS), per-image best threshold (OIS) and average precision (AP), results are summarized in Fig. XXX and Table. 1.
    
    % In the experiment, we first extract ground truth edges on RGB images using Canny detector. Because the motion and illumination condition are mild, RGB images from a standard camera can contain all edges from the environment and be utilized as ground truth. After that, we do RGB-Event image alignment in a coarse-to-fine manner. First, we project the event-edgemap to standard camera image plane by their intrinsic matrices. Then, we further align the two images by Iterative Closest Point (ICP) method. Finally, we measure the edge perception performance using corresponding metrics.
    
    % As shown in Fig. XXX(a), which illustrates the AP regard to the time, the AP of the proposed system is better and more stable than the S-EV. Moreover, in some motion patterns, especially when the moving direction is parallel to most edges in the environment(Fig. XXX(b) and Fig. XXX(c)), the perception from the standard camera would be highly influenced while the one from our system is still stable.
    % The data from the Table. 1 also indicates the same result: our system yields considerable improvements over the S-EV in terms of all metrics. 
    
    \subsection*{Feature Detection and Matching}

 %    \begin{figure*}[htb]
	% \vspace{0.0cm}
	% \centering
	% \includegraphics[width=\textwidth]{figures/Feature-Detection.pdf}
	% \vspace{-0.6cm}
	% \caption{
	% 	The illustration of our feature detection experiments. Fig. (a) is the environment setups of four experiments, figure (b) are the results of our four experiments, on the left half are plots of trackable corner number comparison, on the right is the illustration of gray-scale of event images at a time point/interval, figure (c) is a plot for the comparison on corner detection accuracy and tracking lifetime, figure (d) is a illustration of our improved performance on motion segmentation task.    
	% }
	% \label{fig:featuredet}
 %    \end{figure*}
    
    The following experiments demonstrate the performance of the proposed system for feature detection and matching. These are the most representative tasks in low-level vision and also the basic building blocks for various robotics applications.
    Event-based feature detection and matching are attracting increasing interest\cite{mueggler2017fast, alzugaray2018asynchronous, alzugaray2022event} because of the sensor's advantages due to high dynamic range (HDR) and high temporal resolution. However, the performance of existing methods depends on the camera motion. 
    The proposed system delivers high-quality features independent of camera motion and retains the benefits of event cameras. Suppl. Movie S3 shows the experiments.
    
    %To provide a comprehensive evaluation of the proposed system in this task, the 
    The environments used in the experiments are shown in Fig. \ref{fig:featuredet}A. We used four typical scenarios: a structured environment, an unstructured environment, a challenging illumination environment, and a dynamic environment. The first three scenarios were used for corner-feature detection and tracking, and the last for motion-feature detection and matching, also known as motion segmentation. For all experiments, we compared the proposed system with grayscale cameras and S-EV. We directly extracted features from the asynchronous event stream without any accumulation, preserving the high temporal resolution (in the order of microseconds) of the data.
   % To achieve a denser output stream with higher temporal resolution and less tracking error, the system is expected to maintain compensation accuracy. 
   In these experiments, the wedge angle was set to $1.0^{\circ}$, and the rotating frequency was $12Hz$, which was sufficient to allow for motion compensation at the speed used, as shown in the Suppl. Movie S3 (see the analysis in Materials and Methods). 
    
    \subsubsection*{Corner Detection and Tracking}
    We used the three experimental environments shown in Fig. \ref{fig:featuredet}A.
    After AMI generation, the corner events were extracted using a widely-used event-based corner detector \cite{mueggler2017fast}.
    Next, the extracted features are compensated to eliminate the effect of the wedge rotation. 
    Fig. \ref{fig:featuredet}B shows that our system detected and tracked more corner features and provided more information than S-EV in all three scenarios. The texture in S-EV became unstable due to changing motion, resulting in incomplete corner detection and unstable tracking.
    Furthermore, our system, along with S-EV, outperformed the standard camera in challenging illumination scenarios because of the event sensor's HDR, as shown in Fig. \ref{fig:featuredet}B(iii). 
    The quantitative results presented in Fig. \ref{fig:featuredet}B(iv) demonstrated that our system achieved a substantially longer tracking lifetime than S-EV, although at the cost of slightly reduced accuracy ($\sim 1.5$ pixels). The error in accuracy is primarily due to numerical computations and imperfect clock synchronization introduced during AMI compensation. Therefore, the error is independent of the camera's motion. For a more detailed analysis of this error, please refer to the Choice of Deflector Angle and Rotating Speed section.
    
    %For more information, readers are referred to  the discussion (section \ref{sec:discussion})
   % on Next-step Challenges and Future Work.
    Moreover, our system and the S-EV had a notably higher update rate than standard cameras, which is crucial in high-dynamic scenarios, as shown in Fig. \ref{fig:featuredet}B(v). 
    In conclusion, our system was the only camera system that robustly detected and tracked corner features in all three typical scenarios. The results demonstrated that it effectively solved the corner detection and tracking tasks, especially in challenging illumination scenarios. 
    
    \subsubsection*{Motion Segmentation}       
    The event camera is well suited for segmenting fast-moving objects, and there is already a wide range of applications, including dynamic obstacle avoidance \cite{he2021fast, falanga2020dynamic, zhou2021eventseg} and high-speed counting \cite{bialik2022fast, prophesee_2020}. 
    This experiment aimed to demonstrate that our system and S-EV have a better performance than standard cameras for this task, and the additionally introduced motion in our system does not affect the performance.
    
    The goal of the experiment is to segment independently moving objects from the background. 
    In the experiment, the camera introduced motion in the background while a separately thrown ball moved independently.%in front of the platform, which intersects with the background and introduces occlusion. 
    For motion segmentation on S-EV and AMI-EV, we adapted the methods from \cite{stoffregen2019event} and \cite{falanga2020dynamic}, which can provide per-event segmentation. Specifically, we employed the idea of camera-motion compensation \cite{mitrokhin2018event,parameshwara20210} by maximizing the sharpness of motion-compensated images and detecting moving objects as non-sharp regions using clustering techniques.
    For the standard camera, we applied one of the state-of-the-art methods \cite{rozumnyi2021fmodetect} as our benchmark, which detects fast-moving objects as a truncated distance function to the trajectory by learning from synthetic data.
 
    Comparing results from S-EV and AMI-EV in Fig. \ref{fig:featuredet}C, we see that the introduced motion does not influence the accuracy and robustness of the proposed system in motion segmentation tasks. However, the standard camera suffers from motion blur and low temporal resolution and cannot effectively capture the motion information, thus resulting in poor performance. More details can be seen in Suppl. Movie S3.

    \subsection*{Human Detection and Pose Estimation}
    This experiment demonstrated the potential of applying the proposed system in a popular high-level vision problem -- Human Detection and Pose Estimation.
    Event cameras are particularly well-suited for detection tasks that involve fast motion and have attracted interest in recent years \cite{xu2020eventcap, zou2021eventhpe, zhang4353603neuromophic}. However, previous methods either need the assistance of grayscale images to update the detection\cite{xu2020eventcap} or initialization of the pose estimation \cite{zou2021eventhpe}. Moreover, they do not apply to dynamic environments where the camera moves.
    % while camera motionis very common in robotics applications because the robot is likely to actively adjust its viewpoint for less occlusion. 
    In this experiment, we demonstrated the potential of the proposed AMI mechanism in achieving robust high-speed human motion estimation. To obtain better texture and intensity information, we used images reconstructed from events as the event representation, which have been proven to be robust in different scenarios, including the dynamic one \cite{gallego2020event, rebecq2019high, wang2021deep}. 
    We utilized one of the most popular human pose estimation algorithms, called OpenPifPaf \cite{kreiss2021openpifpaf}, to conduct human detection and pose estimation.

    We evaluated accuracy and robustness using Intersection over Union (IoU) and Percentage of Detected Joints (PDJ). These evaluations are made in relation to the video frame rate, which denotes the number of frames per second that the stand event-to-video algorithm, E2VID \cite{rebecq2019high}, can generate.
    As shown in Fig. \ref{fig:pose}, the AMI-EV demonstrated better performance at different frame rates. 
    When using our system, the frame rate can be configured to be substantially higher than S-EV while maintaining image quality. More details can be found in Suppl. Movie S4.
    
    % \note{higher dynamic range, texture on the body; previous limit: person cannot move}

% % 	轮廓完整性/高帧率
%     \begin{figure}[htb]
%     \centering
%     \includegraphics[width=1\textwidth]{figures/HPE/walk_demo_ori.png}
%     \caption{Walking demo without rotating wedge lens.}
%     \label{fig:walk_wo}
%     \end{figure}
%     \begin{figure}[htb]
%     \centering
%     \includegraphics[width=1\textwidth]{figures/HPE/walk_demo.png}
%     \caption{Walking demo with rotating wedge lens.}
%     \label{fig:walk_w}
%     \end{figure}
%     \begin{figure}[htb]
%     \centering
%     \includegraphics[width=0.7\textwidth]{figures/HPE/shake_hands.pdf}
%     \caption{Waving hands demo w/wo rotating wedge lens.}
%     \label{fig:walk_w}
%     \end{figure}    

    % \subsubsection*{Semantics Segmentation}
	
	% \subsubsection*{Ablation Experiments}
	
    \subsection*{AMI-EV Simulator and Translator}
    \subsubsection*{AMI-EV Simulator}
   To facilitate future research, we also developed a simulator. The code was released in \cite{botao_he_2023_8175198}.
    The simulator was based on our previous work,  WorldGen\cite{Singh2022WorldGenAL}, which allows the generation of 3D photorealistic scenes with the user having control of features, like the scene texture and the camera and lens properties.
    %, for better generalizability by diminishing the data bias in the network.
   The simulator allows the user to generate task-specific synthetic AMI-EV.
    Fig. \ref{fig:simulator}A illustrates an example of a scene created for human pose estimation. The simulator provided the synthetic AMI-EV data along with a list of visual representations of the scene. See the Suppl. Methods for more details on the simulator.
    
\subsubsection*{AMI-EV Translator}
    In addition to the simulator, we also provided a translator to create synthetic AMI-EV from standard datasets. The proposed translator supports three types of inputs: greyscale images, greyscale images combined with events, or events only. With appropriate video interpolation algorithms, high framerate videos can be generated. Subsequently, these high framerate videos are fed into a specially designed AMI module to produce the output AMI event stream. To understand the working principle of the AMI-EV translator in detail, please refer to the Suppl. Methods and Suppl. Fig. S5.
    Fig. \ref{fig:simulator}B shows translation examples from two typical event-based datasets, called Neuromorphic-Caltech101\cite{orchard2015converting} and Multi Vehicle Stereo Event Camera Dataset  \cite{Zhu2018TheMS}, which are both widely used for evaluating event-based 3D perception and recognition tasks. Further results can be found  in the Suppl. Methods.

    \section*{Discussion}
    \label{sec:discussion}
   By emulating
    %mimicking (rather emulating) 
    the biological microsaccade mechanism, a texture-enhancing event vision system, which enables high-quality data association has been proposed and evaluated.  Stable texture appearance and high informational output are maintained with our system consisting of a rotating wedge filter in front of an event camera.
    %anytime by physically adding an active saccade-like motion. 
   % However, processing the new event data format, which encodes both rotational motion and camera motion, requires unconventional algorithms. 
    %To overcome this challenge, we further 
    We also provided a compensation algorithm to account for the motion from the wedge filter.
    %decouples the added motion from the camera motion.   
    Our results show that the output of the compensation is compatible with most representative event-based data processing methods with minimal loss of accuracy and latency. For low-level tasks, there is a margin of error introduced by this compensation, particularly in specialized tasks like optical flow estimation over short time intervals. As shown in Suppl. Fig. S11, the performance of optical flow estimation to static objects degrades by 0.19 pixels of End-Point Error (EPE). However, the benefits of preserving stable texture generally outweigh this drawback. For high-level tasks, the effect of this loss is negligible as it doesn't compromise the performance of advanced recognition or detection tasks.
    % Overall, our proposed texture-enhanced event vision system, along with the accompanying compensation algorithm, represents a significant advancement in the field of event-based vision.

We demonstrated experimentally that our device can acquire more environmental information than traditional S-EVs.
 %Our experiments demonstrate that our system 
 It can maintain a high-informational output while preserving the advantages of event cameras, such as HDR and high temporal resolution. 
  Extensive validation experiments demonstrated that our system has potential for use in various field robotics applications, ranging from low-level to high-level vision tasks. It can achieve better feature extraction in low-level vision tasks, and help the robot recognize and understand the environment better in high-level vision tasks.

    In summary, our proposed system fundamentally eliminates the motion dependency problem  in event-based vision using a bio-inspired mechanism. This hardware improvement enables our system to easily achieve high-quality data output compared to S-EVs. Furthermore, the proposed software allows our system to be used for elaborate mission-specific requirements.
    
    \subsection*{Future Work}
    As shown before, the proposed hardware device and  software solution allow better data association for event-based vision. However, the system is less energy-efficient than an S-EV, because of the additional mechanical structure. What's more, the different data format also calls for additional data processing methods.
	
    To make the hardware more energy-efficient, future research will need to improve the AMI generating mechanism both in the hardware and software.
    Most actuators of this size consume energy from watts to a few tens of watts, which is higher than the S-EV. 
    To achieve less power consumption 
    %comparable to or even much less than that of the S-EV, 
  %  one possible solution from the hardware perspective is to 
    one could replace the mechanical structure with electro-optic materials and control the incoming light direction by Optic Phase Array (OPA)\cite{mcmanamon1996optical} technology. Specifically, by  dynamically controlling the optical properties of electro-optic materials like Liquid Crystal Display (LCD) \cite{schadt1997liquid}, the direction of the incoming light can be steered. 
    % Such approaches can achieve very high control frequency ($>60Hz$ by a SLM (Spatial Light Modulator) \cite{efron1994spatial} and $>5000Hz$ by MEMS (Micro-Electro-Mechanical System) \cite{allen2005micro}) while maintaining low power consumption.
    Such approaches can achieve over 5 KHz control frequency by Micro-Electro-Mechanical System (MEMS) \cite{tilmon2020foveacam} while maintaining low power consumption, which has been validated and applied in the computational imaging field \cite{tilmon2020foveacam, haessig2019spiking}.
    Another possible solution is to optimize the rotation speed and adapt it to the specific scenario. The effect of the added AMI motion decreases with faster scene motion. High-speed rotation is more effective for low-dynamic scenarios; thus, its use could be adapted to the speed.
    For certain tasks, the system could operate at low speed or even stop once an adequate amount of data has been collected for analysis or increase  its rotational speed in response to diminishing texture.
    However, designing specific action strategies for different application scenarios remains a challenge.
    
    The proposed device also creates an event data format where a periodic motion is encoded into the event stream. 
    This raises a question: Is there a more efficient and effective way to process the new data than compensating for it? 
    In this work, the compensation algorithm removes the added motion from the output stream to make it compatible with existing event-based algorithms. However, this method also introduces some discretization errors and adds computational costs. Although the error (around 1.5 to 2.0 pixels) is acceptable for most robotics applications, and the system can still work in real-time on onboard computers, the additional error and computation may be problematic in applications where precise measurements are needed or for small robots with limited computation resources. Moreover, the current compensation procedure amalgamates the events of both polarities and thus loses the polarity features.
    Future work can consider a more complex fitting model, such as an oriented ellipse, instead of a circular motion to further decrease the compensation error.
    To eliminate the compensation error fundamentally, we may need a method that can work directly on the generated event stream and utilize the motion information without moving the pixel locations. 
    {We will also investigate training a neural network to regress the accurate pixel-wise compensations function. In the spirit of event-based work, we could train a spiking neural network. We believe the best way would be to train the network as a regressor using the method of conversion; for example, we first train an Artificial Neural Network (ANN) and then convert this network into an Spiking Neural Network (SNN) \cite{perez2013mapping,diehl2016conversion}.

    \section*{Materials and Methods}

    \subsection*{Hardware Architecture}
    \label{sec:hardware}

 %    \begin{figure*}[ht]
	% \vspace{0.0cm}
	% \centering
	% \includegraphics[width=1.0\textwidth]{figures/system_overview.pdf}
	% \vspace{-0.6cm}
	% \caption{
	% 	\textbf{The overview of our entire system design including both hardware design and software solution.} \textbf{(A)} includes the real-world hardware design and Computer-Aided Design(CAD) model of our hardware design. \textbf{(B)} illustrates a sequence of refraction results of incoming light as the wedge prism rotates. \textbf{(C)} is the AMI generation and compensation process, the image on the left is the accumulation image of the event stream on the right. \textbf{(D)} presents the working process of our entire system.
	% }
	% \label{fig:system}
 %    \end{figure*}

    % Humans use extraocular muscles to randomly change the gaze direction to generate microsaccades in all directions, in our system, a similar but more effective solution is designed to generate such omnidirectional stimuli. 
    % Instead of moving the whole perception system, e.g., eyes, like humans, which is energy-consuming, we utilize the working principle of the rotating wedge-prism deflector\cite{senderakova2003analysis} and design a mechanical device to actively adjust the direction of the incoming light, as is demonstrated in Fig. \ref{fig:system}A and Fig. \ref{fig:system}B. 
    % system: 322
    % platform: 1627
    % event camera: 131
    % act: 57

    The proposed hardware platform is of size
    %system is a 
    $82 \times 54 \times 62$-mm platform. 
   % The total weight of the platform 
    Its total weight is 322 g, including a 131 g event camera (with lens) and a 41 g external MCU for actuator control. 
    Our system comprises four modules: the Optical deflector module, the Actuator module, the Event camera module, and the Mirco Computing Unit (MCU), as shown in Fig. \ref{fig:system}A. The blueprints have been
    %that we designed and assembled, which is also
    released \cite{botao_he_2023_8175198} to benefit future research. 
    
    For the optical deflector module, a wedge prism was mounted in front of the camera lens to deflect the incoming light at a fixed angle 
   from $x_w$, the x-axis in a coordinate frame
   %where $x_w, y_w, z_w$ denote the corresponding axis of the wedge-prism
   $w$ attached to the wedge prism
   %\in \mathbb{S}^2$ 
   (shown in Fig. \ref{fig:system}B).
    The actuator module drove the optical deflector module rotating around $z_c$, the z-axis of the coordinate frame attached to the camera frame $c$, also shown in Fig. \ref{fig:system}B. Our platform uses a  DJI M2006 Brushless DC Motor \cite{m2006} with a customized reduction gear and an absolute position encoder. With the modified gear module, the actuator weighted 57 g (including the electronic speed controller) and provided $0.11 N \cdot m$ torque at 1500 rpm, which satisfied our rotation speed and torque requirements. Moreover, by adding a photoelectric sensor to sense the prism's orientation, the motor's incremental encoder signal could be transformed to get an absolute orientation measurement as needed for the AMI calibration.
% the built-in incremental encoder, we transformed it into an absolute rotary encoder to satisfy the requirement for AMI calibration.
    For the camera module, we adopted the DVXplorer event camera \cite{DVXplorer}. It has a spatial resolution of $640 \times 480$ and supports time synchronization with external sensors.
    The Micro Controller Unit (MCU) was used to control the actuator's motion, receive position feedback, and synchronize the event camera with the actuator.
   % timestamps between the event camera and the actuator's position feedback. 
   %For developing convenience, we adopted 
   We used the DJI Robomaster Development Board \cite{rmboard}, whose weight is 40 g to simplify the development.
    
    \subsection*{Choice of deflector angle and rotation speed}
    In this section, we experimentally evaluated the influence of the rotation speed and prism angle on the data volume and compensation accuracy, and subsequently, we discuss good choices
    %optimal combinations 
    for different tasks.
    As demonstrated in Fig. \ref{fig:deg_spd}, increasing the degree of tilted angle of the wedge prism and rotation speed led to generating a larger number of events but also higher motion compensation errors. 
    
    Two factors governed the selection of rotational speed: the duration of the maintained time window and the compensation error.
   Considering the former, the data must originate from at least a quarter of the rotation period, as this is the smallest unit containing a pair of orthogonal motions
   %. Such motions are crucial 
   necessary for activating edges in all directions and thereby ensuring texture stability. For instance, an event count image typically comprises data spanning a $33ms$ duration. Consequently, one rotation period should last $33 \times 4 = 132 ms$ (~455 rpm) to guarantee the inclusion of all environmental information within a single frame. In practice, the rotational speed must surpass this minimum requirement to counteract the influence of sensor noise.
   
    The second issue was the compensation error.  As illustrated in the left sub-figure of Fig. \ref{fig:deg_spd}B, the error — represented by the standard deviation of the event distribution — surged dramatically beyond $720rpm$ for $0.5^{\circ}$ and $1.0^{\circ}$ prisms. This escalating influence can be attributed to small synchronization errors among the sensors, which amplify as the rotational speed increases. Furthermore, this effect bears a connection to the deflector angles.
    In light of the above analysis, the rotational speed was set to $720rpm$ for all real-world experiments to achieve a balance between texture stability and compensation accuracy.

    The selection of the deflector angle was task-specific. As shown in Fig. \ref{fig:deg_spd}A, the geometric structure was similar across the output of all three tested prisms, with the primary differences being data volume and compensation accuracy.
    For tasks that prioritize data intensity, such as corner detection and tracking, a larger tilt angle was preferable, provided that accuracy was maintained. This is because a prism with a larger tilt angle can generate more events in a given time, as shown in Fig. \ref{fig:deg_spd}B, and these events are mostly found in areas with rich texture features, such as corner points. This leads to increased robustness in such tasks.
    Conversely, a smaller tilt angle was more suitable for tasks emphasizing contour completeness and compensation accuracy as long as data sufficiency was ensured. For instance, in tasks like human pose estimation or semantic segmentation, the completeness and sharpness of object boundaries are more critical than the data intensity.
    According to the left sub-figure of Fig. \ref{fig:deg_spd}B, both the $0.5^{\circ}$ and $1.0^{\circ}$ prisms exhibited satisfactory compensation accuracy at a rotational speed of $720 rpm$.
    In the right sub-figure, the $1.0^{\circ}$ prism displayed higher data intensity than the $0.5^{\circ}$ one. The $2.0^{\circ}$ prism, although it had the highest data intensity, its compensation error was too high to be practical.
    Therefore, in this work, we chose a $1.0^{\circ}$ prism with the rotation speed at $720 rpm$ for the feature detection and matching experiments and a $0.5^{\circ}$ prism with $720 rpm$ in all other experiments, as well as in the simulator and translator.

    % It's worth noting that the rate of error increase is more gradual for a prism tilted angle of $0.5^{\circ}$ compared to tilted angles of $1.0^{\circ}$ and $2.0^{\circ}$, as shown in the right subplot of Fig. \ref{fig:deg_spd}. This is largely attributed to the complex interplay between the lens tilted angle and rotation speed, which is difficult to model in the motion compensation algorithm.
    
    \subsection*{Microsaccade Model and its Simplification}
    \subsubsection*{2-D Wedge-prism Camera Model.}
    %To provide an intuitive example of how the optical model of our proposed system works,
    Fig. \ref{fig:methodology}A illustrates the optical model with a 2-D cross-section of the wedge-prism camera model. 
    %Due to the reversibility of the optical path~\cite{XXX}, we reverse the analysis of the optical path from the photosensitive sensor. 
    The incoming light $v_{in} \in \mathbb{S}^1$ denoting a unit vector on the left, was transmitted and deflected twice ($v'$ and $v_{out}$) through the wedge prism and then focused on the camera image plane $I$ at pixel $p_i$.
    According to Snell's Law \cite{born2013principles}, the relationship between $\Phi_i$ and $\Phi_p$
    % $\textbf{v_{in}} and \textbf{v^{'}}$, $\textbf{v^{'}} and \textbf{v_{out}}$
    can be described as: 
    \begin{equation}
    \begin{aligned}
    sin \Phi_i &= n \cdot sin \Phi_p \\
    \Phi_p &= arcsin\left(\frac{sin \Phi_i}{n}\right),
    \end{aligned}
    \label{eq:snell-2d}
    \end{equation}
  where $\Phi_i$ 
    is the angle between $\mathbf{v_{in}}$ and $\mathbf{z_c}$, and $\Phi_p$ is the angle between $\mathbf{v_{p}}$ and $\mathbf{z_c}$, respectively. $n$ is the refractive index of the prism material, which was set to $1.55$ in the experiments. The refractive index of the air is regarded as $1.0$.
    Therefore, vector $\mathbf{v_{p}}$ can be represented as:
    \begin{equation}
    \mathbf{v_{p}} = R\left(\widehat{\mathbf{v_{i}} \times \mathbf{z_c}}, \Phi_i - \Phi_p\right) \cdot \mathbf{v_{i}},
    \end{equation}
    where $R(a, b)$ denotes a rotation along axis $a$ with an angle of $b$ (anti-clockwise as the positive direction), and $\widehat{\mathbf{v}}$ denotes the normalized vector $\mathbf{v}$.

    Similarly, the relationship between $\Phi_q$ (angle between $\mathbf{v_{p}}$ and $\mathbf{z_w}$) and $\Phi_o$ (angle between $\mathbf{v_{o}}$ and $\mathbf{z_w}$) can be written as
    \begin{equation}
    \Phi_o = arcsin\left(n \cdot sin \Phi_q\right),
    \end{equation}
    where $\Phi_q$ can be expressed as
    \begin{equation}
    \Phi_q = arcsin\left(||\mathbf{v_p} \times \mathbf{z_w}||\right).
    \end{equation}
    Finally, the output light vector $\mathbf{\mathbf{v_o}}$ can be represented as:
    \begin{equation}
    \mathbf{v_{o}} = R\left(\widehat{\mathbf{v_{p}} \times \mathbf{z_w}}, \Phi_q - \Phi_o\right) \cdot \mathbf{v_{p}},
    \end{equation}    

   Summarizing, $\mathbf{\mathbf{v_o}}$ can be fully represented by $\mathbf{v_i}$ and $\mathbf{z_w}$. The transmission through the prism is described by a function $g(\mathbf{v_i}, z_c) \in \mathbb{S}^1$ as:
    \begin{equation}
    g(\mathbf{v_i}, \mathbf{z_w}) = R(\widehat{\mathbf{v_{p}} \times \mathbf{z_w}}, \Phi_q - \Phi_o) \cdot R(\widehat{\mathbf{v_{i}} \times \mathbf{z_c}}, \Phi_i - \Phi_p).
    \label{eq:g1}
    \end{equation}
    Since  $\mathbf{v_{p}}, \mathbf{z_w}, \mathbf{v_{i}}, \mathbf{z_c}$ are in the same plane, $\widehat{\mathbf{v_{p}} \times \mathbf{z_w}}, \widehat{\mathbf{v_{i}} \times \mathbf{z_c}}, \widehat{\mathbf{z_w} \times \mathbf{z_c}}$ are parallel to each other.
    According to the pinhole camera model \cite{renner2012pinhole}, the angle between $\mathbf{v_{p}}$ and $\mathbf{z_w}$, $\mathbf{v_{i}}$ and $\mathbf{z_c}$ are larger than $90^{\circ}$, which means $\widehat{\mathbf{v_{i}} \times \mathbf{z_c}} = \widehat{\mathbf{z_c} \times \mathbf{z_w}}$ and $\widehat{\mathbf{v_{p}} \times \mathbf{z_w}} = \widehat{\mathbf{z_w} \times \mathbf{z_c}}$. Therefore, $g(\mathbf{v_i}, z_c)$ can also be written as:
    \begin{equation}
    \begin{aligned}
    g(\mathbf{v_i}, \mathbf{z_c}) &= R(\widehat{\mathbf{z_w} \times \mathbf{z_c}}, \Phi_q - \Phi_o) \cdot R(\widehat{\mathbf{z_c} \times \mathbf{z_w}}, \Phi_i - \Phi_p) \\
    &= R(\widehat{\mathbf{z_w} \times \mathbf{z_c}}, \Phi_q - \Phi_o) \cdot R(\widehat{\mathbf{z_w} \times \mathbf{z_c}}, \Phi_p - \Phi_i) \\
    &= R(\widehat{\mathbf{z_w} \times \mathbf{z_c}}, \Phi_q - \Phi_o + \Phi_p - \Phi_i) \\
    &= R(\widehat{\mathbf{z_w} \times \mathbf{z_c}}, \delta(\mathbf{v_i, z_w}))
    \end{aligned}
    \label{eq:g2}
    \end{equation}
    where $\delta(\mathbf{v_i, z_w}) = \Phi_q - \Phi_o + \Phi_p - \Phi_i$ because these variables are determined by $\mathbf{v_i}$ and $\mathbf{z_w}$ according to Snell's Law \cite{born2013principles}.

    Eventually, based on the pinhole camera model \cite{renner2012pinhole}, $\mathbf{\mathbf{v_o}}$ can be projected to the image plane by the camera's intrinsic matrix $K$, and the wedge-prism camera model can be formulated as:
    \begin{equation}
        p_i = \mathbf{K} \cdot g(\mathbf{v_i, z_w}) \cdot \mathbf{v_i}
    \end{equation}

    \subsubsection*{Rotating Wedge-prism Camera Model}
  Building on the 2-D wedge-prism camera model, we next explain the 3-D rotating model. 
	In Fig. \ref{fig:methodology}B, the incoming light $\mathbf{v_{i}} \in \mathbb{S}^2$, is transmitted and deflected twice ($\mathbf{v_p} \in \mathbb{S}^2$ and $\mathbf{v_{o}} \in \mathbb{S}^2$) through the wedge prism, and finally focused by the lens on the image plane, at $I_{m, n}$, where m and n are the indexes of the image pixel.
	
 %   Different from the 2-D model, 
    The rotating wedge-prism camera model introduces a time-varying rotation, which adds a variable $\theta$, as shown in Fig. \ref{fig:methodology}B.
   Therefore, the transmission from $\mathbf{v_{i}}$ to $\mathbf{v_{o}}$ is defined as $G(\mathbf{v_{i}}, \mathbf{z_w(\theta)})$ generalizing
 %   The format of $G(\mathbf{v_{i}}, \mathbf{z_w(\theta)})$ is similar to 
    $g(\mathbf{v_{i}}, \mathbf{z_w})$  in Eq. \ref{eq:g1} with  a parameter for time  $t$ added because $\mathbf{z_w(\theta)}$ is time-varying. The transmission function $G(\mathbf{v_{i}}, \mathbf{z_w(\theta)})$ can be expressed as:
    
    \begin{equation}
        G(\mathbf{v_{i}}, \mathbf{z_w(\theta)}) = R(\mathbf{v_{p}} \times \mathbf{\mathbf{z_w(\theta)}}, \Phi_q - \Phi_o) R(\mathbf{v_{i}} \times \mathbf{z_c}, \Phi_i - \Phi_p).
    \label{eq:G}
    \end{equation}
    
    Thus, the transmission from $\mathbf{v_{i}}$ to $\mathbf{v_{o}}$ can be expressed as:
    
    \begin{equation}
        \mathbf{v_{o}} = G(\mathbf{v_{i}}, \mathbf{z_w(\theta)}) \cdot \mathbf{v_{i}}.
    \label{eq:v_o}
    \end{equation}

    Finally, $\mathbf{\mathbf{v_o}}$ can be projected onto the image plane, and the camera's intrinsic matrix is denoted as $K$. The proposed rotating wedge-prism camera model can be formulated as:
    \begin{equation}
        I(m,n) = \mathbf{K} \cdot G(\mathbf{v_{i}}, \mathbf{z_w(\theta)}) \cdot \mathbf{v_{i}}
    \end{equation}

    \subsubsection*{Microsaccade Model Simplification}
    % \note{Need to think more about the motivation of simplification, it should be simplified because the original model is time-varying and do not have analytical solution}
    With the proposed optical model, the optical properties of our system can be precisely described. However, its accuracy is highly dependent on the spatial resolution of $\mathbf{v_{i}}$ and $\theta$. The resolution is negatively related to the robustness of the calibration. For example, for a $640 \times 480$ event camera, if the resolution $\theta$ is set as $1^{\circ}$, it needs $640 \times 480 \times 360$ parameters to fully describe the model. If so, the calibration process needs a long time to collect enough data for each pixel, and any illumination change during the process will highly influence the results. If we down-sample the resolution, a discretization error will be introduced, resulting in poor compensation performance. 

    To make the parameter calibration more efficient in memory and computation, we simplified the model and reduced the number of parameters by applying an approximation. 
    Firstly, we decomposed $\mathbf{v_i}$ into two vectors $\mathbf{v_{\perp}}$ and $\mathbf{v_{\parallel}}$, where $\mathbf{v_{\perp}}$ is vertical to $\mathbf{z_{w}(\theta)} \times \mathbf{z_{c}}$ and $\mathbf{v_{\parallel}}$ is parallel to $\mathbf{z_{w}(\theta)} \times \mathbf{z_{c}}$. These two vectors can be expressed as:
    \begin{equation}
    \begin{aligned}
    \mathbf{v_{\parallel}} &= (\mathbf{v_i} \cdot \widehat{\mathbf{z_{w}(\theta)} \times \mathbf{z_{c}}}) \cdot \widehat{(\mathbf{z_{w}(\theta)} \times \mathbf{z_{c}})}, \\
    \mathbf{v_{\perp}} &= \mathbf{v_i} - \mathbf{v_{\parallel}},
    \end{aligned}
    % \label{eq:snell-2d}
    \end{equation}
    where $\widehat{\mathbf{z_{w}(\theta)} \times \mathbf{z_{c}}}$ is the normalized unit vector of $\mathbf{z_{w}(\theta)} \times \mathbf{z_{c}}$.
    Then Eq. \ref{eq:v_o} can be written  as:     

    \begin{equation}
    \begin{aligned}
        \mathbf{v_{o}} &= G(\mathbf{v_{\perp}} + \mathbf{v_{\parallel}}, \mathbf{z_w(\theta)}) \cdot \mathbf{v_{i}} \\
        &\approx g(\widehat{\mathbf{v_{\perp}}}, \mathbf{z_w(\theta)}) \cdot \mathbf{v_{i}} \\
        &= g(\widehat{\mathbf{v_{\perp}}}, \mathbf{z_w(\theta)}) \cdot \mathbf{v_{\perp}} + g(\widehat{\mathbf{v_{\perp}}}, \mathbf{z_w(\theta)}) \cdot \mathbf{v_{\parallel}} \\
        &= \left(g(\widehat{\mathbf{v_{\perp}}}, \mathbf{z_w(\theta)}) \cdot \widehat{\mathbf{v_{\perp}}}\right) \cdot ||\mathbf{v_{\perp}}|| + \mathbf{v_{\parallel}},
    \end{aligned}
    \label{eq:simp_v_o}
    \end{equation}  
    where from  Eq. \ref{eq:g2}, we have that $g(\widehat{\mathbf{v_{\perp}}}, \mathbf{z_w(\theta)}) = R(\widehat{\mathbf{z_c} \times \mathbf{\mathbf{z_w(\theta)}}}, \delta(\mathbf{v_i, \mathbf{z_w(\theta)}}))$ .

%    Till now, with the change of $t$, 
    The trajectory of $\mathbf{v_i}$ over time is shown in Fig. \ref{fig:methodology}B. It is close to, but not exactly, a circle in $SO(3)$.
  %  but not exactly. 
    This is because the rotation axis $\mathbf{z_c}$ is not aligned with $\mathbf{\mathbf{z_w(\theta)}}$, resulting in the change of $||\mathbf{v_i \times \mathbf{z_w(\theta)}}||$. 
	Therefore, the radius $\delta(\mathbf{v_i, \mathbf{z_w(\theta)}})$ also varies over time, and we denote the set of $\delta(\mathbf{v_i, \mathbf{z_w(\theta)}})$ as $\Delta = \delta^i(i = 1, 2...)$. 
 
    Still, since $\Delta$ has hundreds of parameters, further simplification is needed. Thus, we defined a new frame $w'$ that is fixed to the wedge prism and rotates with it. $w'$ has the same origin as $w$, and their Z-axes $z_{w'}$ and $z_w$ are aligned.
 % , the transformation can be expressed as:
	% \begin{equation}
	%     w' = R(X_W, -\alpha) \cdot W.
	% \end{equation}	
    Since $\mathbf{z_{w'}}$ is aligned with the rotational axis $z_c$, $||\mathbf{v_i \times z_w'}||$ is constant.
    Now, $g(\widehat{\mathbf{v_{\perp}}}, \mathbf{z_w(\theta)})$ can be represented as:
    \begin{equation}
    \begin{aligned}
        g(\widehat{\mathbf{v_{\perp}}}, \mathbf{z_w(\theta)}) &= R(\widehat{\mathbf{z_c} \times \mathbf{\mathbf{z_w(\theta)}}}, \delta(\mathbf{v_i, z_w'})) \\
        &= R(\widehat{\mathbf{z_c} \times \mathbf{\mathbf{z_w(\theta)}}}, \delta(\mathbf{v_i, z_c})).
    \end{aligned}
    \end{equation}
 %    Then, we replace the $w$ with $w'$, the new model can be denoted as:
	% \begin{equation}
	% \begin{cases}
 %    \rho_1 = arccos(\frac{\mathbf{v_3}-(\mathbf{v_3} \cdot \mathbf{X_{w'}}) \cdot \mathbf{Z_{w'}}}{\lVert \mathbf{v_3}-(\mathbf{v_3} \cdot \mathbf{X_{w'}}) \cdot \mathbf{Z_{w'}} \rVert} \cdot \mathbf{Z_{w'}})\\
 %    \rho_2^{'} = arcsin(sin \alpha \sqrt{n^2 - sin^2\rho_1} - cos\alpha \cdot \sin\rho_1) \\
 %    \delta'(\mathbf{v_1}) = \rho_2^{'} + \rho_1 - \alpha \\
 %    f(I_{m,n}) = g'(\mathbf{v_3}, \theta) = R(\mathbf{X_{w'}}, \delta'(\mathbf{v_3})) \\
 %    \end{cases}.
	% \end{equation}
% 	\begin{equation}
%         \delta'(\mathbf{v_1}) = \rho_2^{'} + \rho_1 - \alpha,
%     \end{equation}
%     \begin{equation}
%         f(I_{m,n}) = g'(\mathbf{v_3}, \theta) = R(\mathbf{X_{w'}}, \delta'(\mathbf{v_3})).
%     \end{equation}
    In this model, as shown in Fig. \ref{fig:methodology}C, $\mathbf{\mathbf{z_w(\theta)}} = R(\mathbf{z_c}, \theta) \cdot R(\mathbf{X_C}, \alpha) \cdot \mathbf{z_c}$, and $\theta(t) = \theta_b + \tilde{\theta}(t)$, where $\theta_b$ denotes the bias of initial position between the actuator encoder and the circular trajectory and $\tilde{\theta}(t)$ denotes the angular measurement obtained from the encoder.

    Through the above approximation, the trajectory of $\mathbf{v_i}$ is simplified to a circle $\odot \phi(\delta, \theta) \in SO(3)$, which brings two main advantages. First, it only has two parameters, $\delta$ and $\theta$, for each pixel, which are easy to calculate, store, and optimize. Second, it is differentiable, which means it does not lose accuracy due to discretization. Admittedly, this simplification also introduces some errors. Our analysis found that the error is within 2 pixels in a $90^\circ$ FOV, which can be safely ignored. Details of the analysis are in Suppl. Methods.
    % In practice, because of the parameter estimation error and the numerical error during the compensation procedure, the final error is unlikely to be as small as theoritical analysis. However, according to the experimenal result of \note{xxx}, 
    % \begin{figure}[htb]
    % \centering
    % \includegraphics[width=0.6\textwidth]{figures/Simplification.pdf}
    % \caption{(a) Original model. (b) Simplified model.}
    % \label{fig:simplification}
    % \end{figure}
    
    % https://aip.scitation.org/doi/full/10.1063/1.5011979?casa_token=0z14F0c_eZMAAAAA:jrMPjnCGXuc-4VRH1as07Nsawxvt6kBvcK6FAzutkQBypqbgqxLd4vTwlKNze6Y_H3GzWhplgMUF

    \subsection*{Microsaccade Calibration and Compensation}

    The calibration procedure calibrates $\delta$ and $\theta_{b}$ in an optimization-based manner. The first step of our algorithm is to assign the initial values for $\delta$ and $\theta_b$, denoted as $\delta^0$ and $\theta_b^0$. The choice of initial values was based on the hardware setup. $\delta^0$ was set as the refraction angle of the wedge-prism, and the zero-position of the encoder determined $\theta_b^0$.
    In our procedure, we collected a batch of events $E = e_i(i = 1, 2...)$ and encoder data over some time $t$ -- $t = 2s$. In the experiment, we found that this achieved a good balance between calibration accuracy and computational cost. There was not enough information for shorter time windows, and sometimes it did not converge. Large time windows could result in a heavy burden on the computation because millions of events had to be processed in each iteration. Still, they did not increase the accuracy notably because, with 2 seconds, there were already $15$ or more periods of rotation, which was sufficient for the computation. 
    Next, we transferred the events from the spatial-temporal domain $(x,y,t)$ to the $(x,y,\theta)$ domain by synchronizing the events' timestamp with the wedge-prism's angular position. 
    Then, we warped all events back to $\theta_0$ to compensate for the rotation.

	The warping function is described as $\Pi: \mathbb{R}^3 \xrightarrow{} \mathbb{R}^3$, which warps the event's position on image plane as $\Pi(x, y, \tilde{\theta}(t)): (x, y, \theta) \xrightarrow{} (x', y', \theta_0)$. The warping function can be written as:
	\begin{equation}
	    e_i' = \Pi\{(x, y, \theta)\} = \mathbf{K} \cdot g'^{-1}(\mathbf{v_1}, \mathbf{\mathbf{z_w(\theta)}}) \cdot \mathbf{K}^{-1} \cdot
	    \begin{bmatrix}
	        x \\
	        y \\
	        1 
	    \end{bmatrix}
	    = (x', y', \theta_0).
	\end{equation}
    From the warped events $E' = e'_i(i = 1, 2...)$, we constructed the Image of Wrapped Events (IWE) \cite{gallego2018unifying} $H$ as:
	\begin{equation}
        H = \sum_{e_i \in E'} \zeta(e_i'),
	\end{equation}
    where each pixel $(i,j)$ sums the warped events $e_i'$ that mapped to it. $\zeta$ represents intensity spikes, where $\zeta(e_i') = 1$ means $e_i'$ is mapped to $(i,j)$, otherwise $\zeta(e_i') = 0$.
    To evaluate the quality of this calibration parameter pair, we designed a cost function by leveraging the idea of motion compensation \cite{stoffregen2019event}.
    Since a well-parameterized IWE will warp events triggered by the same incoming light to the same pixel, the IWE should be sharp. Therefore, we designed our cost function $J$ to measure the sharpness of the IWE:
    
    \begin{equation}
        % J = \sum_{i,j} \frac{1}{1+e^{h'(i,j)}}
        J = \sum_{i,j} b_{i,j}\left(1+exp\left(\frac{h(i,j)}{\eta}\right)\right)^{-1},
    \end{equation}
    where $h(i,j)$ is the value of pixel $(i,j)$ in $H$, and $\eta$ is the scale factor. If $h(i,j)$ is positive, then $b_{i,j}$ is set to 1; otherwise, $b_{i,j} = 0$ so that the cost would not be summed. 
    We used the exponential in the above equation, 
    as it heavily weighted pixels with low numbers of events.
    Therefore, the sharpness of IWE is inversely proportional to the cost, as shown in Fig. \ref{fig:methodology}C.
    The optimal parameter pair $\delta, \theta_0$ was optimized by maximizing the sharpness, or contrast of IWE:  $\min_{\delta, \theta_0} J$.

    In practice, the above equation was robustly solved by a coarse-to-fine search. The search process was demonstrated in Fig. \ref{fig:methodology}C. It was formulated as a standard circular function fitting problem as shown in Eq. S1. The Suppl. Fig. S3 indicates that the optimal solution is unique. Moreover, in Suppl. Methods, we further prove that Eq. S1 is convex in a certain domain, which means it can be solved much faster if the initial guess is precise.
	
	% \begin{figure}[htb]
 %    \centering
 %    \includegraphics[width=0.6\textwidth]{figures/Calib_demo.pdf}
 %    \caption{Demonstration of }
 %    \label{fig:simplification}
 %    \end{figure}
	
    \section*{Supplementary Materials}
    % Sect. S1. Data collection platform.\\ 
    % Sect. S2. Perceptual fading effect in event cameras.\\
    % Sect. S3. Convexity analysis of the fitting error in AMI calibration.\\
    % Sect. S4. Quantitative error analysis of the AMI generation simplification.\\
    % Sect. S5. Overview of the released simulator and translator\\
    % Sect. S6. Discussion about the effect of AMI on moving objects.\\
    % Sect. S7. Details about compensation error measurement.\\
    % Sect. S8. Analysis of bandwidth-related noise.\\
    % Fig. S1. Overview of the data collection platform.\\
    % Fig. S2. Demonstration of perceptual fading effect in event cameras.\\
    % Fig. S3. Illustration of the fitting error in AMI calibration.\\
    % Fig. S4. Illustration of the error caused by AMI calibration and compensation.\\
    % Fig. S5. Overview of the proposed simulator and translator.\\
    % Fig. S6. An illustration of the results produced by the proposed simulator.\\
    % Fig. S7. Coupling effect of the AMI motion and camera motion.\\
    % Fig. S8. Entropy of the accumulated event image.\\
    % Fig. S9. Calculation of the compensation error.\\
    % Fig. S10. Comparison of event density distribution.\\
    % Fig. S11. Comparison of optical flow estimation results.\\
    % Fig. S12. Demonstration of data transmission delay.\\
    % Fig. S13. Bandwidth comparison between S-EV and AMI-EV.\\
    % Movie S1. Demonstration of Microsaccade.\\
    % Movie S2. AMI generation and compensation mechanism.\\
    % Movie S3. Enhancement of texture in different event representations\\
    % Movie S4. Feature detection and matching. \\
    % Movie S5. Human detection and pose estimation.\\

    Supplementary Methods. \\
    Supplementary Figures S1 to S13. \\
    Supplementary Movies S1 to S4.

\phantom{\cite{tulyakov2021time, hu2021v2e, gehrig2020video, jiang2018super, shi2015convolutional, nuc, cpu, DenizRFB23, tulyakov2022time, gehrig2021raft}}

\bibliography{scibib}
\bibliographystyle{Science}

\section*{Acknowledgments}
We thank Xin Zhou, Levi Burner, and Snehesh Shrestha for their valuable suggestions for the manuscript. 
We sincerely appreciate the work of Junxiao Lin, Long Xu, and Zhihao Tian for their help in real-world experiments. \\
Furthermore, we are truly grateful for Dr. Yi Zhou's suggestions regarding the design of the experiment. \\
\textbf{Funding:} This work was supported by the National Natural Science Foundation of China under grant No. 62322314, the National Science Foundation of the U.S. under grant OISE 2020624
supporting Research through International Network-to-Network Collaboration ("AccelNet: Accelerating Research on Neuromorphic Perception, Action, and Cognition"), and the National Natural Science Foundation of China under grant No. 62088101. \\
\textbf{Author Contributions:} \\
Conceptualization: BH, FG, CF \\
Methodology: BH, ZW, FG, CF, YA \\
Investigation: BH, ZW, YZ, JC, CDS, HL \\
Visualization: BH, YG, ZW, HL, YZ, CDS \\
Funding Acquisition: FG, CF, YA \\
Project Administration: FG, CF, YA, CX, YC, KW \\
Supervision: FG, CF, YA, CX, YC, KW, SS \\
Writing – Original Draft: BH, ZW, CF, FG, YA \\
Writing – Review \& Editing: BH, ZW, CF, FG, YA \\
\textbf{Competing Interests:} The authors declare they have no competing interests. \\
\textbf{Data and materials availability:} Data needed to evaluate the conclusions in the paper have been made available for download at \url{https://zenodo.org/records/8157775} \cite{botao_he_2023_8175198}. \\

    \section*{Movies and figures}
	
    Movie 1: \textbf{Demonstration of microsaccades and overview of the proposed system.}\\
    \begin{figure}[ht]
    \centering
    \includegraphics[width=0.8\textwidth]{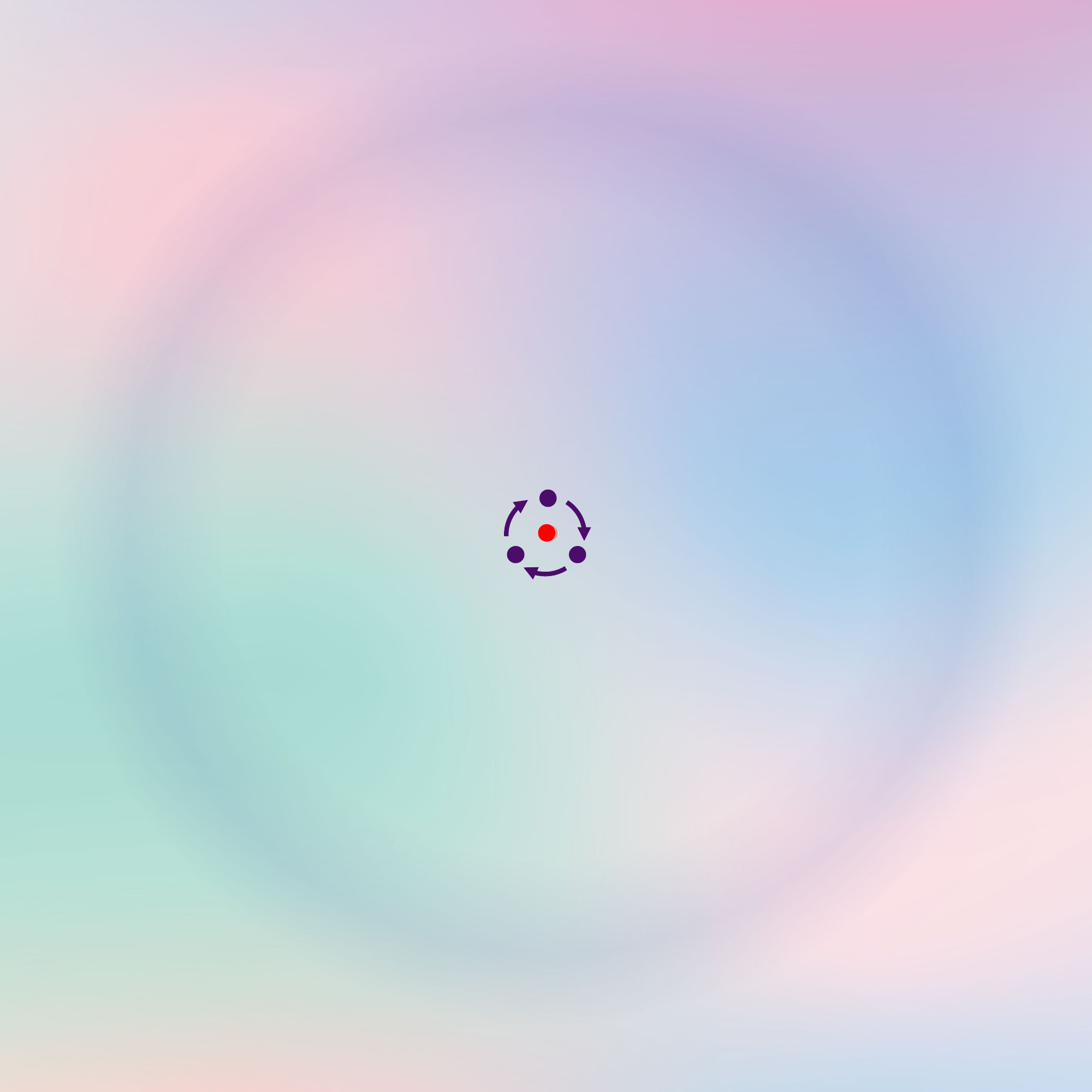}
    \caption{\textbf{Demonstration of how microsaccades counteract visual fading.} A simple yet intuitive example demonstrating visual fading and how microsaccades counteract it. We recommend enlarging the image to at least $15 \times 15$ cm and keeping your eyes 40cm away from the screen. After a few seconds of fixation on the red spot, the bluish annulus and the background will fade. This is because microsaccades are suppressed during this time, and therefore, the eye cannot provide effective visual stimulation to prevent peripheral fading. On the other hand,  when saccading between the purple spots, the annulus is always experienced, possibly fading slower even though the saccades are small, typically $0.5^\circ$-$1.0^\circ$ depending on the viewer's distance from the figure.}
    \label{fig:ms}
    \end{figure}

    % \begin{figure}[htb]
    % \centering
    % \includegraphics[width=0.6\textwidth]{figures/fig2.png}
    % \caption{Illustration for losing shape.}
    % \label{fig:illu_shape}
    % \end{figure}

    \begin{figure*}[ht]
    \vspace{0.0cm}
    \centering
    \includegraphics[width=1.0\textwidth]{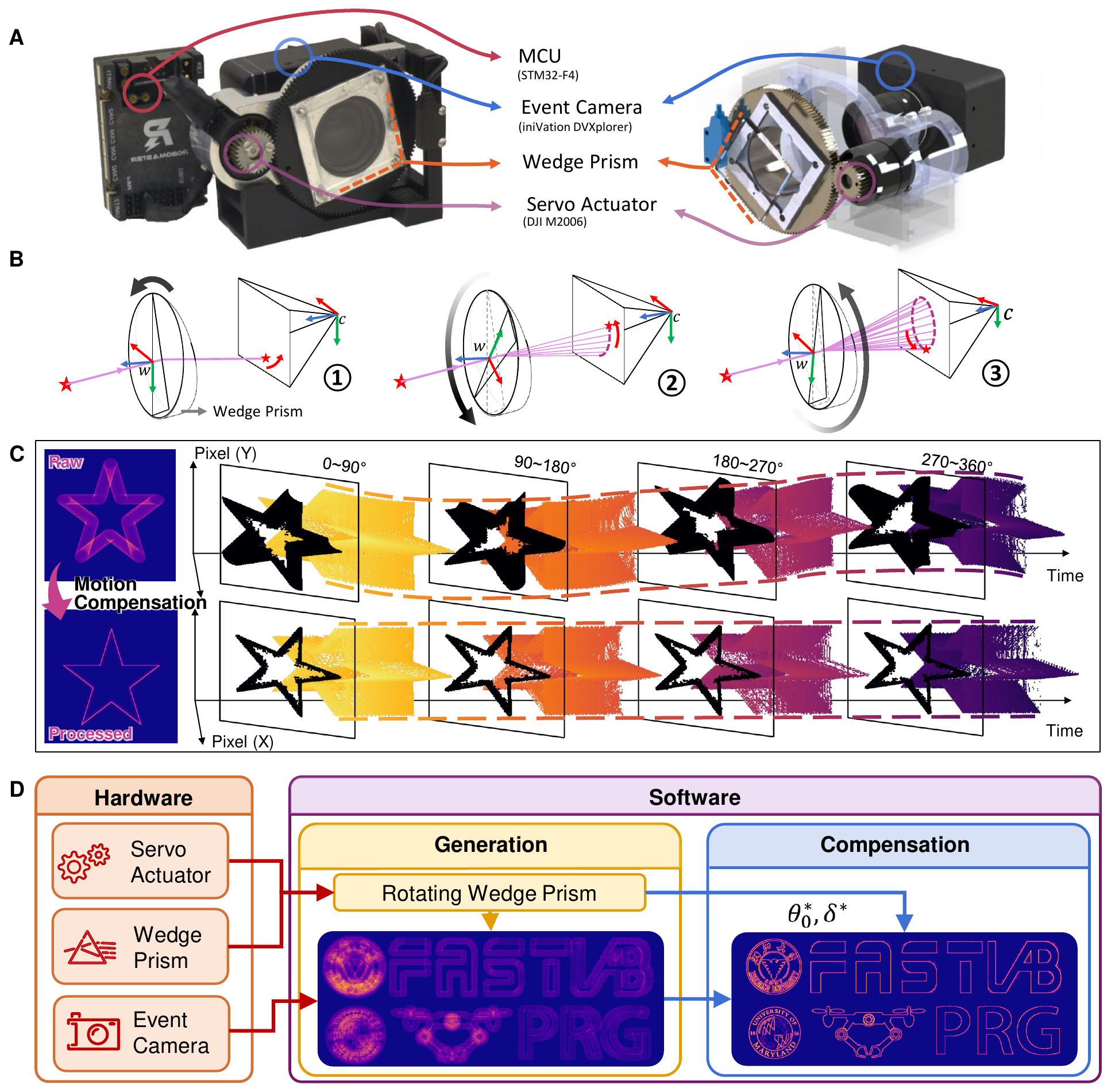}
    \vspace{-0.6cm}
    \caption{
    \textbf{Overview of our entire system, including both hardware and software.} \textbf{(A)} Real-world hardware and Computer-Aided Design (CAD) model. \textbf{(B)} Illustration of the incoming light refraction as the wedge prism rotates. \textbf{(C)} Event generation and compensation process, with the images on the left resulting from accumulating the event streams shown on the right. \textbf{(D)} System overview.
    }
    \label{fig:system}
    \end{figure*}

    \begin{figure*}[htb]
	\vspace{-0.5cm}
	\centering
	\includegraphics[width=1.0\textwidth]{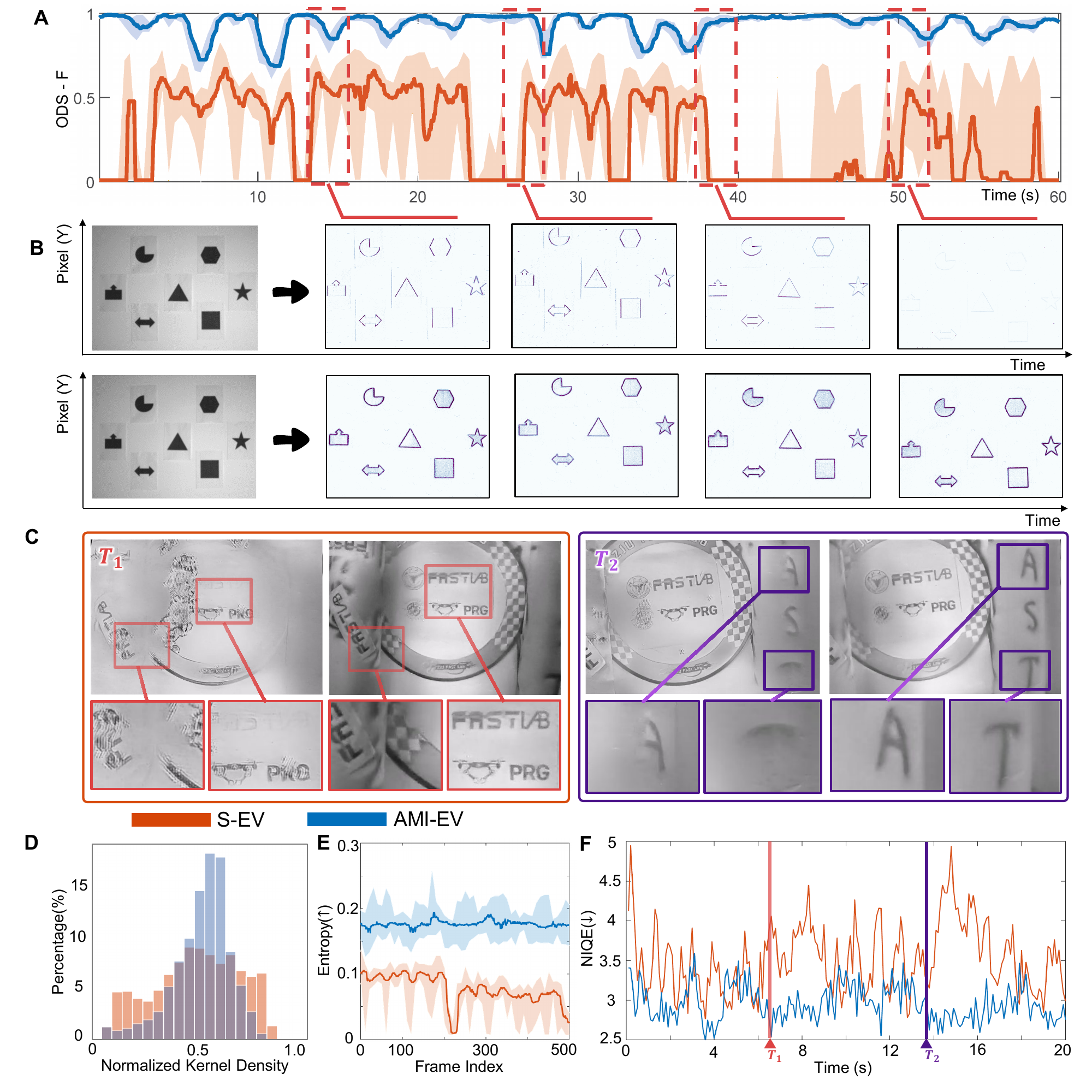}
	\vspace{-0.7cm}
	\caption{
		\textbf{Illustration of our approach's improvement on texture enhancement.} \textbf{(A)} The ODS-F (higher is better) is used to measure the structural completeness of the accumulated event images.  \textbf{(B)} Temporal snapshots of (A). \textbf{(C)} Comparison of the reconstructed gray-scale images. (C) is the snapshots of (F), the color red for the box is used to indicate that the system is static, and purple denotes that the system is moving upward (along Y-axis). 
        \textbf{(D)} Histogram of Event Density Distribution for the original event stream and our enhanced event stream. More detailed illustrations can be found in Suppl. Fig. S10.
        \textbf{(E)} Entropy comparison of accumulated event images. In (A) and (E), solid curves indicate the median value over a time window of 10 data points. In contrast, the top and bottom bounds of the transparent regions indicate their maximum and minimum values.
        \textbf{(F)} Quantitative comparison of the reconstructed image quality using the Natural image quality evaluator (NIQE, lower is better) \cite{mittal2012making}.
	}
	\label{fig:texture}
    \end{figure*}

    \begin{figure*}[htb]
	\vspace{0.0cm}
	\centering
	\includegraphics[width=\textwidth]{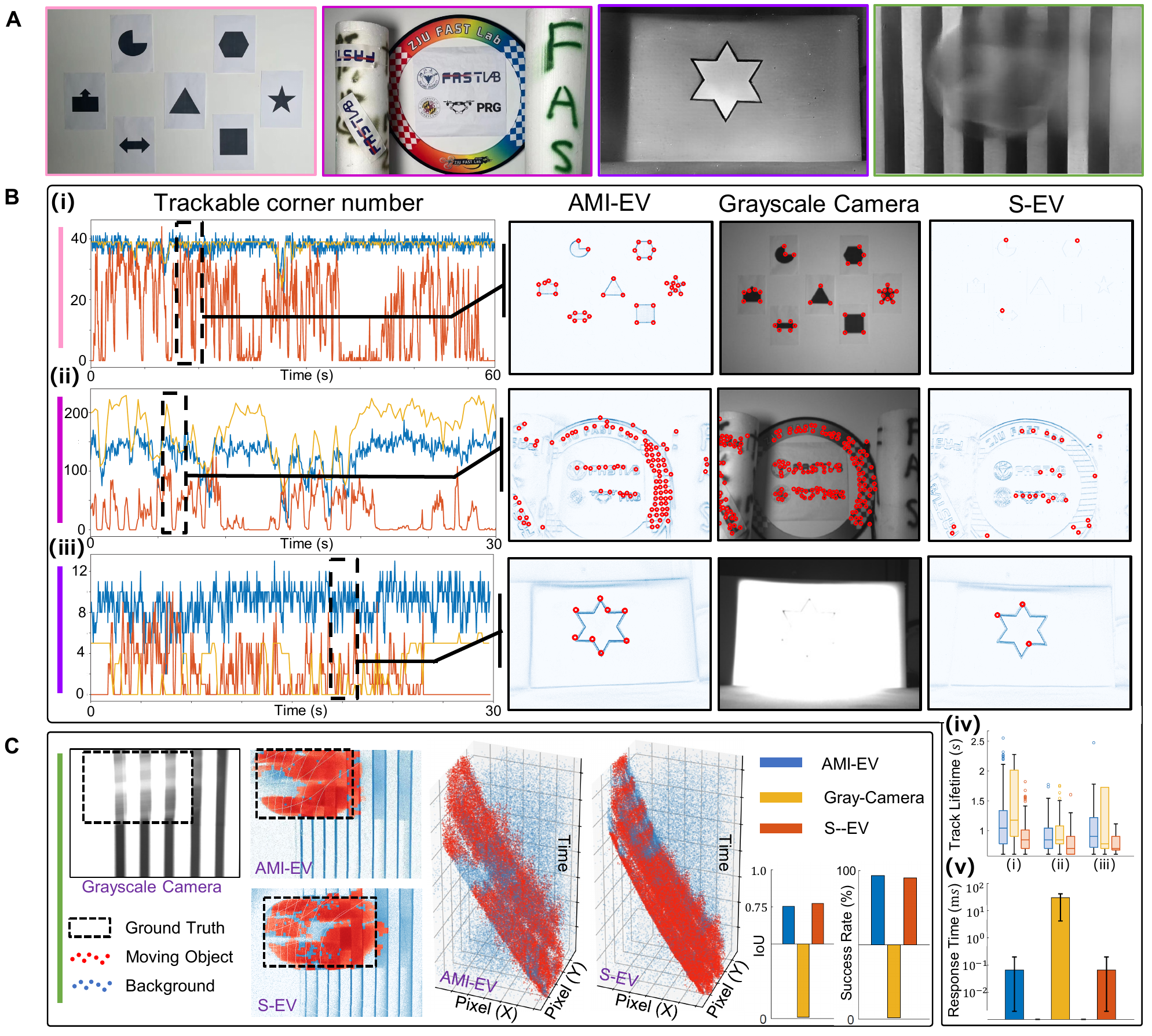}
	\vspace{-0.6cm}
	\caption{
        \textbf{Evaluation of feature detection and matching.} \textbf{(A)} Environment setups of four experiments. \textbf{(B)} Results of the corner detection and tracking experiments. The left column of (i)-(iii) provides a comparison of the number of trackable corners, and the three right columns show snapshots. (iv) and (v) are metric comparisons visualized using box and bar graphs. (iv) indicates the lifetime of all trackable corners, and (v) shows the response time. 
        \textbf{(C)} Results of the motion segmentation experiment. Blue parts indicate the background and red parts indicate independently moving objects.
	}
	\label{fig:featuredet}
    \end{figure*}

    \begin{figure}[htb]
    \centering
    \includegraphics[width=\textwidth]{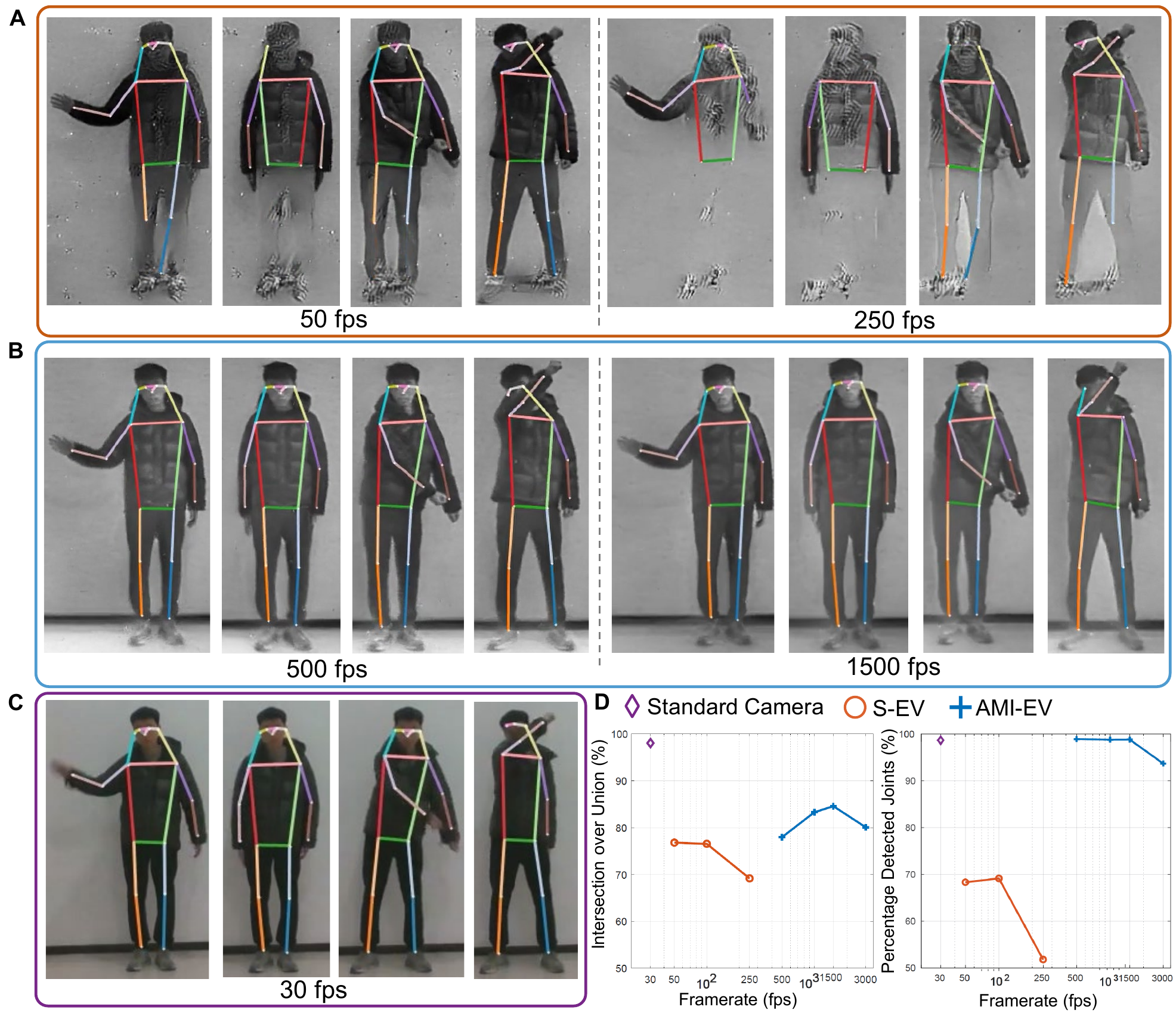}
    \caption{
    \textbf{Evaluation of human detection and pose estimation.} 
    \textbf{(A-C)} Results of human pose estimation for S-EV (A), AMI-EV (B) and a standard camera (C) on four actions: wave the hand, shake arms, baseball batting action, and ping-pong batting. The former two actions are slow and the latter two are fast. 
    \textbf{(D)} Metrics comparisons. The framerate denotes the number of frames per second that the standard event-to-video algorithm, called E2VID \cite{rebecq2019high}, is configured to generate. Intersection over Union (IoU) provides a measure of human detection performance, and Percentage Detected Joints (PDJ) is a measure of the detected joints' localization precision and completeness. Because the sampling frame rate varies greatly from different sensors, we use the Semilog plot (x-axis has log scale) to visualize the data.
    }
    \label{fig:pose}
    \end{figure}  

    \begin{figure}[htb]
    \centering
    \includegraphics[width=\textwidth]{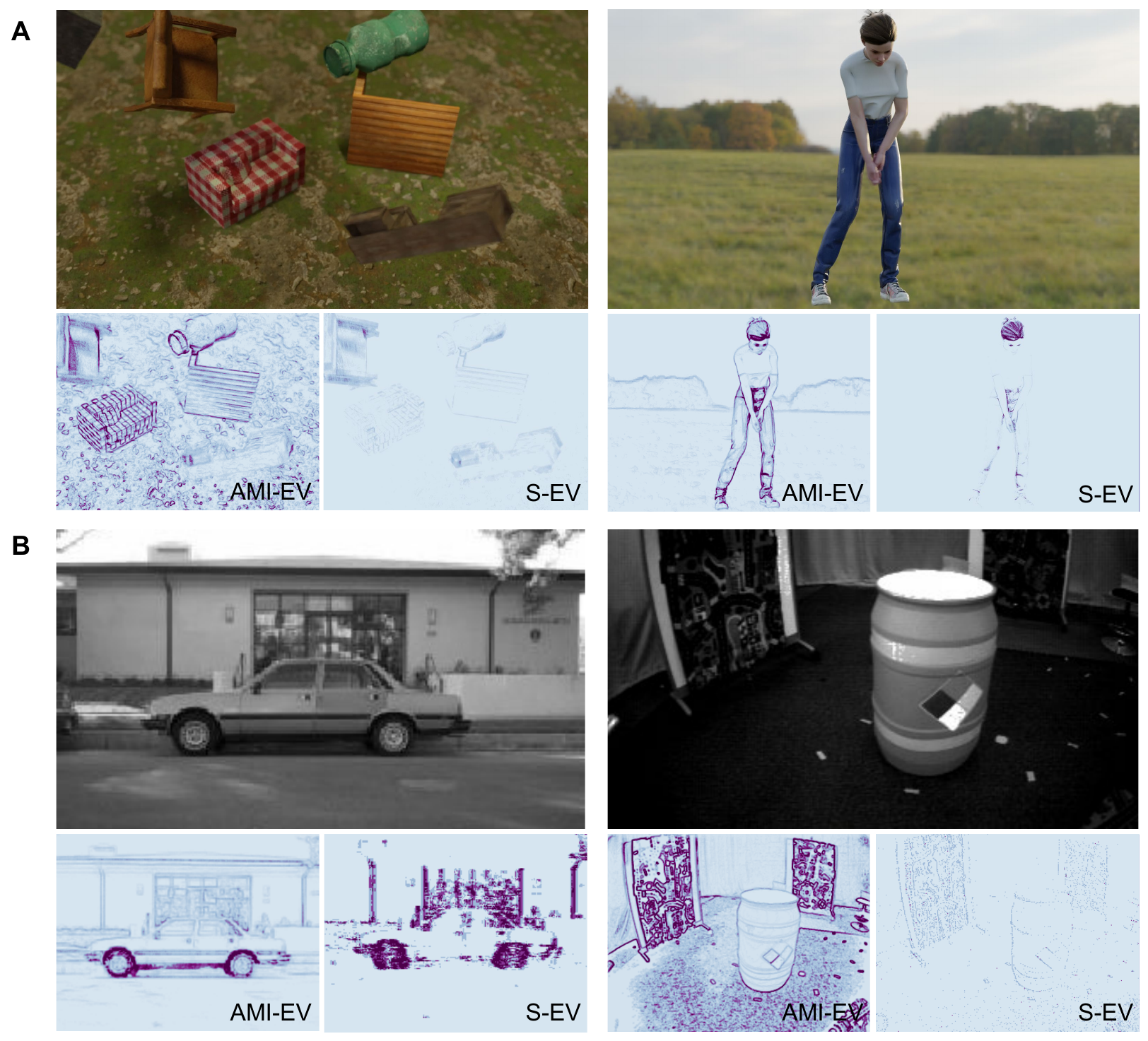}
    \caption{
    \textbf{Pictures generated by the released software package.}
    \textbf{(A)} (left) 3D rendered scene with multiple moving objects (right) golf scene.
    \textbf{(B)} Output of the released translator. (left) image from the Neuromorphic-Caltech 101 dataset and two event count images generated from an S-EV and an AMI-EV, respectively; (right) scene from Multi Vehicle Stereo Event Camera Dataset \cite{zhu2018multivehicle}.
    }
    \label{fig:simulator}
    \end{figure}

    \begin{figure}[htb]
    \centering
    \includegraphics[width=\textwidth]{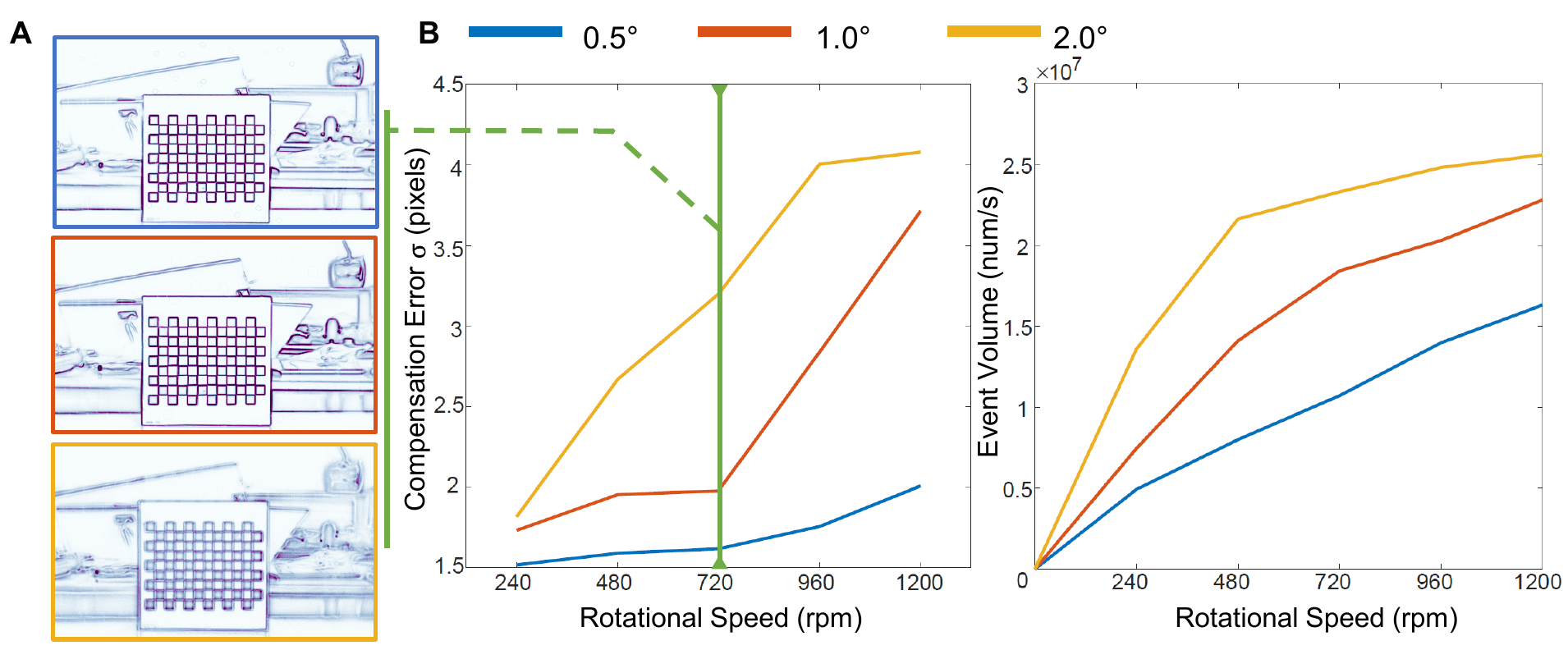}
    \caption{\textbf{Compensation error and data volume for different combinations of deflector angles and rotational speeds.} 
    \textbf{(A)} A snapshot of compensation performance with the rotational speed at 720 rpm. The colors on the image boundaries indicate the deflector angles.
    \textbf{(B)} Quantitative results. Details about how to calculate the compensation error can be found in Suppl. Methods. The event volume is measured by bandwidth analysis, as detailed in Suppl. Fig. S9.
    } 
    \label{fig:deg_spd}
    \end{figure}

    \begin{figure}[htb]
    \centering
    \includegraphics[width=1\textwidth]{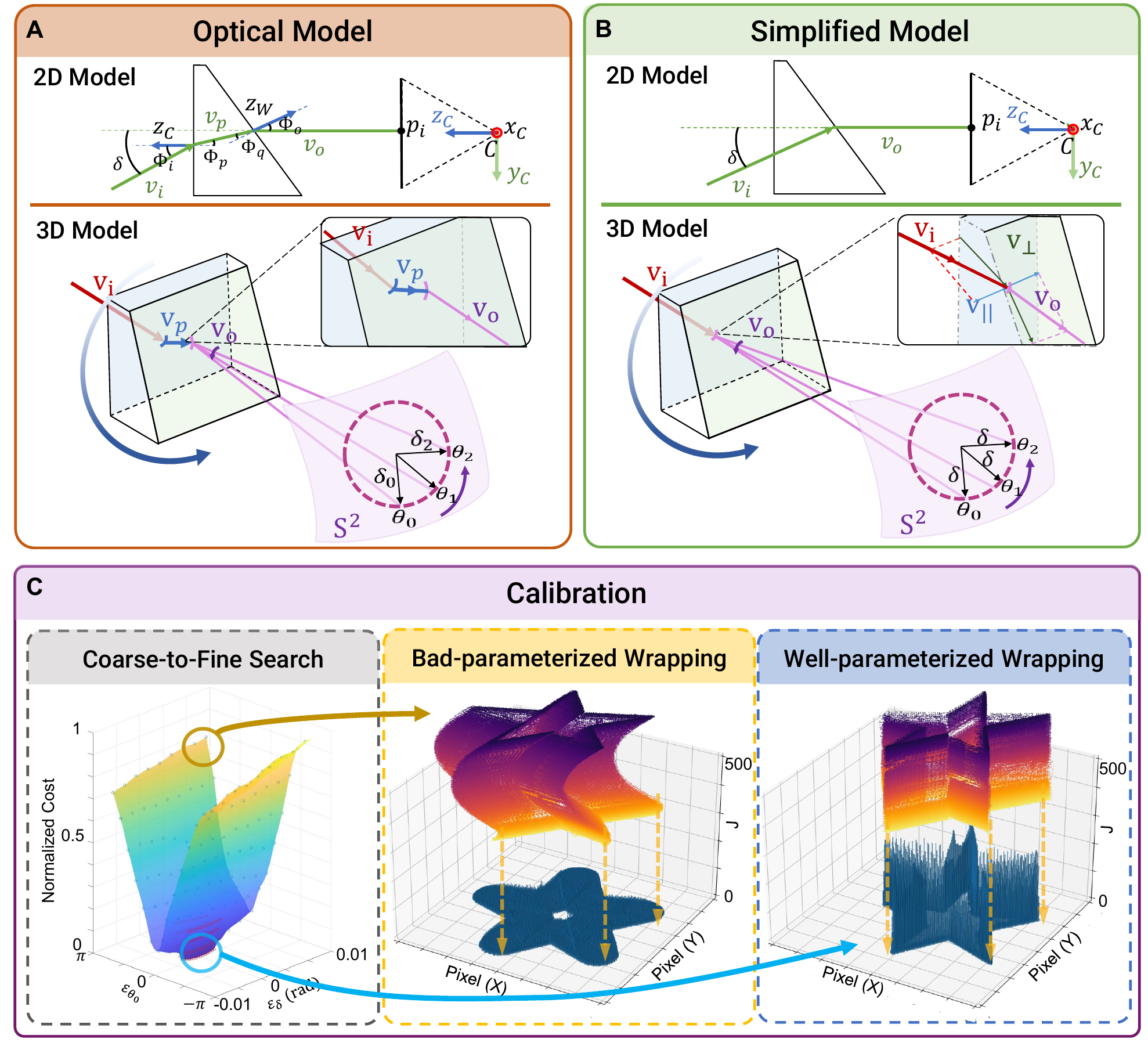}
    \caption{\textbf{Demonstration of the optical model, model simplification, and calibration procedures of our system.} \textbf{(A)} The 2-D wedge-prism camera optical model and the 3-D rotating model of the proposed AMI-EV. \textbf{(B)} A simplified model of (A). \textbf{(C)} The calibration procedure. (left) The  Coarse-to-fine search procedure, where blue points are samples from coarse search, and red points are samples from fine search (bottom of the surface). A surface was fitted to the samples. (middle) Bad estimate of the actual trajectory, with a sharpness cost of $29,569$. (right) Good estimate of the actual trajectory, with a sharpness cost of $2,382$.
}
    \label{fig:methodology}
    \end{figure}

\end{document}